\documentclass[graybox,12pt,oneside,a4paper]{svmult}
\pdfoutput=1
\usepackage[bottom]{footmisc}
\usepackage{hyperref}       
\usepackage{longtable}
\usepackage{pifont,type1cm,graphicx,multicol,newtxtext,newtxmath,xcolor,enumitem,cancel,bm,bbm,etoolbox,algorithm,algorithmic,dsfont,tikz,wrapfig,subfigure}
\usepackage{xcolor}
\usepackage[algo2e]{algorithm2e}
\tikzstyle{ocont}=[ellipse,draw=black!100,thick,minimum size=6mm,>=stealth]  
\tikzstyle{dgraph}=[->, line width=1.5pt]
\usetikzlibrary{arrows,shapes}
\newcommand\Tstrut{\rule{0pt}{3.6ex}}         
\newcommand\Bstrut{\rule[-1.9ex]{0pt}{0pt}}   
\newcommand{\tc}{\mathcal{C}}
\newcommand{\W}{\mathcal{W}}
\newcommand{\independent}{\rotatebox[origin=c]{90}{$\models$}}

\def\1{\mathbbm{1}}
\newcommand{\trans}{^{\scriptscriptstyle \top}}

\DeclareMathOperator*{\argmin}{arg\,min}

\usepackage[textwidth=3cm,textsize=scriptsize]{todonotes}

\makeindex
\usepackage[top=2cm, bottom=2cm, left=3cm, right=3cm]{geometry}
\begin{document}
\title*{Fairness in Machine Learning}
\author{Luca Oneto and Silvia Chiappa}
\institute{
Luca Oneto \at University of Pisa, Pisa, IT
\email{luca.oneto@unipi.it}
\and
Silvia Chiappa \at DeepMind, London, UK
\email{csilvia@google.com}
\and
In: Oneto L., Navarin N., Sperduti A., Anguita D. (eds) Recent Trends in Learning From Data. Studies in Computational Intelligence, vol 896. Springer, Cham, 2020. \href{https://doi.org/10.1007/978-3-030-43883-8_7}{doi.org/10.1007/978-3-030-43883-8{\_}7}.
}
\maketitle
\abstract{
Machine learning based systems are reaching society at large and in many aspects of everyday life.
This phenomenon has been accompanied by concerns about the ethical issues that may arise from the adoption of these technologies.
ML fairness is a recently established area of machine learning that studies how 
to ensure that biases in the data and model inaccuracies do not lead to models that treat individuals unfavorably on the basis of characteristics such as e.g. race, gender, disabilities, and sexual or political orientation.
In this manuscript, we discuss some of the limitations present in the current reasoning about fairness and in methods that deal with it, and describe some work done by the authors to address them. More specifically, we show how causal Bayesian networks can play an important role to reason about and deal with fairness, especially in complex unfairness scenarios. We describe how optimal transport theory can be used to develop methods that impose constraints on the full shapes of distributions corresponding to different sensitive attributes, overcoming the limitation of most approaches that approximate fairness desiderata by imposing constraints on the lower order moments or other functions of those distributions. We present a unified framework that encompasses methods that can deal with different settings and fairness criteria, and that enjoys strong theoretical guarantees. We introduce an approach to learn fair representations that can generalize to unseen tasks. Finally, we describe a technique that accounts for legal restrictions about the use of sensitive attributes.
}
\section{Introduction}
\label{sec:Introduction}
Machine Learning (ML) is increasingly used in a wide range of decision-making scenarios that have serious implications for individuals and society, including financial lending~\cite{byanjankar2015predicting,malekipirbazari2015risk}, hiring~\cite{bogen2018hiring,hoffman2018discretion}, online advertising~\cite{he2014practical,perlich2014machine}, pretrial and immigration detention~\cite{angwin2016machine,rosenberg2018immigration}, child maltreatment screening~\cite{chouldechova2018child,vaithianathan2013child}, health care~\cite{defauw2018clinically,kourou2015machine}, social services~\cite{ainowinstitute2018litigating,eubanks2018automating}, and education~\cite{oneto2019learning,oneto2017dropout,papamitsiou2014learning}.
Whilst algorithmic decision making can overcome undesirable aspects of human decision making, 
biases in the training data and model inaccuracies can lead to decisions that treat individuals unfavorably (\emph{unfairly}) on the basis of characteristics such as e.g. race, gender, disabilities, and sexual or political orientation (\emph{sensitive attributes}).

ML fairness is an emerging area of machine learning that studies how to ensure that the outputs of a model do not depend on sensitive attributes in a way that is considered unfair. For example, in a model that predicts student performance based on previous school records, this could mean ensuring that the decisions do not depend on gender. Or, in a model that decides whether a person should be offered a loan based on previous credit card scores, this could mean ensuring that the decisions do not depend on race. 

Over the last few years, researchers have introduced a rich set of definitions formalizing different fairness desiderata that can be used for evaluating and designing ML systems. Many such definitions express properties of the model outputs with respect to the sensitive attributes. However, in order to properly deal with fairness issues present in the training data, relationships among other relevant variables in the data need to be accounted for. Lack of consideration for the specific patterns of unfairness underlying the training data has lead e.g. to the use of inappropriate fairness criteria in the design and evaluation of the COMPAS pretrial risk assessment tool. This problem was highlighted in \cite{chiappa2019causal}, by using Causal Bayesian Networks (CBNs) as a simple and intuitive visual tool to describe different possible data-unfairness scenarios. CBNs can also be used as a powerful quantitative tool to measure unfairness in a dataset and to help researchers develop techniques for addressing it. As such, CBNs represent an invaluable framework to formalize, measure, and deal with fairness.

From a procedural viewpoint, methods for imposing fairness can roughly be divided into three families. Methods in the first family consist in pre-processing or extracting representations from the data to remove undesired biases. The pre-processed data or extracted representations can then be used as input to a standard ML model. 
Methods in the second family consist in enforcing a model to produce fair outputs through imposing fairness constraints into the learning mechanism. Methods in the third family consist in post-processing the outputs of a model in order to make them fair. 
Most methods (as well as most fairness definitions) approximate fairness desiderata through requirements on the lower order moments or other functions of distributions corresponding to different sensitive attributes. Whilst facilitating model design, not imposing constraints on the full shapes of relevant distributions can be restrictive and problematic.
Also, most often the goal of these methods is to create a fair model from scratch on a specific task.
However, in a large number of real world applications using the same model or part of it over different tasks might be desirable. To ensure that fairness properties generalize to multiple tasks, it is necessary to consider the learning problem in a multitask/lifelong learning framework.
Finally, there still exist only a few attempts to group methods that can deal with different settings and fairness criteria into a unified framework accompanied by theoretical guarantees about their fairness properties.

In this manuscript, we introduce the problem of fairness in machine learning and describe some techniques for addressing it. We focus the presentation on the issues outlined above, and describe some approaches introduced by the authors to address them. More specifically, in Sect. \ref{sec:CBNs} we stress the crucial role that CBNs should play to reason about and deal with fairness, especially in complex unfairness scenarios. In Sect. \ref{sec:CDOT} we introduce a simple post-processing method that uses optimal transport theory to impose constraints on the full shapes of distributions corresponding to different sensitive attributes. In Sect. \ref{sec:GFERM} we describe a unified fairness framework that enjoys strong theoretical guarantees. In Sect. \ref{sec:LFR} we introduce a method to learn fair representations that can generalize to unseen tasks. In Sect. \ref{sec:FairModels:Sens} we discuss legal restrictions with the use of sensitive attributes, and introduce an in-processing method that does not require them when using the model. Finally, we draw some conclusions in Sect. \ref{sec:Conclusions}. In order to not over-complicate the notation, we use the most suited notation in each section.
\section{Causal Bayesian Networks: An Essential Tool for Fairness}
\label{sec:CBNs}
Over the last few years, researchers have introduced a rich set of definitions formalizing different fairness desiderata that can be used for evaluating and designing ML systems \cite{adler2018auditing,ali2019loss,bechavod2018Penalizing,berk2017convex,calders2009building,calders2013controlling,calders2010three,chiappa2019path,chierichetti2017fair,chouldechova17fair,chzhen2019leveraging,corbett-davies2017algorithmic,donini2018empirical,dwork12fairness,dwork2018decoupled,feldman2015certifying,fish2015fair,fish2016confidence,fitzsimons2018equality,fukuchi2015prediction,goh2016satisfying,hajian2011rule,hardt2016equality,hashimoto2018fairness,he2014practical,hebert2017calibration,heidari2018fairness,heidari2018moral,jabbari2017fairness,jiang2019wasserstein,johnson2016impartial,joseph2016fairness,kamiran2009classifying,kamishima2012fairness,kamishima2013independence,kearns2018preventing,kilbertus2017avoiding,kim2018fairness,komiyama2017two,komiyama2018nonconvex,kusner2017counterfactual,lum2016statistical,luo2015discrimination,luong2011k,mancuhan2012discriminatory,mary2019fairness,mcnamara2017provably,olfat2017spectral,oneto2019taking,oneto2019general,pedreshi2008discrimination,perez-suay2017fair,pleiss2017fairness,raff2018fair,russell2017worlds,speicher2018unified,wang2018invariant,williamson2019fairness,woodworth2017learning,wu2016using,yang2017measuring,yao2017beyond,yona2018probably,zafar2017fairness,zafar2017fairnessARXIV,zafar2017parity,zehlike2017matching,zhang2017causal,zhang2018achieving,vzliobaite2011handling}. There exist several surveys that give an overview of current definitions and describe their relations (see e.g. \cite{gajane2017formalizing,mitchell2018fair,verma2018fairness}). In this manuscript, we are interested in highlighting the risk in the use of definitions which only express properties of the model outputs with respect to the sensitive attributes. As the main sources of unfairness in ML models are biases in the training data, and since biases can arise in different forms depending on how variables relate, accounting for relationships in the data is essential for properly evaluating models for fairness and for designing fair models. This issue was pointed out in~\cite{chiappa2019causal}, by showing that Equal False Positive/Negative Rates and Predictive Parity were inappropriate fairness criteria for the design and evaluation of the COMPAS pretrial risk assessment tool. In Sect. \ref{sec:CBNs:VT}, we re-explain the issue, and also show that making a parallel between Predictive Parity and the Outcome Test used to detect discrimination in human decision making suffers from the same problem of not accounting for relationships in the data. We do that by using Causal Bayesian Networks (CBNs)~\cite{dawid2007fundamentals,pearl2000causality,pearl2016causal,peters2017elements,spirtes2000causation}, which currently represent the simplest and most intuitive tool for describing relationships among random variables, and therefore different possible unfairness scenarios underlying a dataset. In Sect. \ref{sec:CBNs:QT}, we show that CBNs also provide us with a powerful quantitative tool to measure unfairness in a dataset and to develop techniques for addressing it.
\subsection{Causal Bayesian Networks: A Visual Tool for (Un)fairness}
\label{sec:CBNs:VT}
For simplicity of exposition, we restrict ourselves to the problem of learning a binary classification model from a dataset $\mathcal{D} =\{(a^n,x^n,y^n)\}_{n=1}^N$, where each datapoint $(a^n,x^n,y^n)$---commonly corresponding to an individual---contains a binary outcome $y^n$ that we wish to predict, a binary attribute $a^n$ which is considered sensitive, and a vector of features $x^n$ to be used, together with $a^n$, to form a prediction $\hat y^n$ of $y^n$. 

We formulate classification as the task of estimating the probability distribution $\mathbb{P}(Y|A,X)$, where\footnote{Throughout the manuscript, we denote random variables with capital letters, and their values with small letters.} $A, X$ and $Y$ are the random variables corresponding to $a^n,x^n$, and $y^n$ respectively, and assume that the model outputs an estimate $s^n$ of the probability that individual $n$ belongs to class 1, $\mathbb{P}(Y=1|A=a^n,X=x^n)$. A prediction $\hat y^n$ of $y^n$ is then obtained as
$\hat y^n=\1_{s^n>\tau}$, where $\1_{s^n>\tau}=1 \text{ if } s^n>\tau$ for a threshold $\tau\in[0,1]$, and zero otherwise.\\
Arguably, the three most popular fairness criteria used to design and evaluate classification models are \emph{Demographic Parity}, \emph{Equal False Positive/Negative Rates}, and \emph{Calibration/Predictive Parity} \cite{hardt2016equality}.\\

\noindent {\bf Demographic Parity.} Demographic Parity requires $\hat Y$ to be statistically independent of $A$ (denoted with $\hat Y \independent A $), i.e.
\begin{align}
\mathbb{P}(\hat Y=1|A=0)=\mathbb{P}(\hat Y=1|A=1)\,.
\end{align}
This criterion was recently extended into \emph{Strong Demographic Parity} \cite{jiang2019wasserstein}. Strong Demographic Parity considers the full shape of the distribution of the random variable $S$ representing the model output by requiring 
$S\independent A$. This ensures that the class prediction does not depend on the sensitive attribute regardless of the value of the threshold $\tau$ used.\\

\noindent {\bf Equal False Positive/Negative Rates (EFPRs/EFNRs).} EFPRs and EFNRs require 
\begin{align}
&\mathbb{P}(\hat Y=1|Y=0,A=0)=\mathbb{P}(\hat Y=1|Y=0,A=1),\nonumber \\ 
&\mathbb{P}(\hat Y=0|Y=1,A=0)=\mathbb{P}(\hat Y=0|Y=1,A=1).
\end{align}
These two conditions can also be summarized as the requirement $\hat Y \independent A | Y$, often called \emph{Equalized Odds}.\\

\noindent {\bf Predictive Parity/Calibration.} 
Calibration requires $Y \independent A | \hat Y$. In the case of continuous model output $s^n$ considered here, this condition is often instead called \emph{Predictive Parity} at threshold $\tau$, $\mathbb{P}(Y=1|S>\tau,A=0)=\mathbb{P}(Y=1|S>\tau,A=1)$, and Calibration defined as requiring $Y \independent A |S$.\\

\noindent Demographic and Predictive Parity are considered the ML decision making equivalents of, respectively, Benchmarking and the Outcome Test used for testing discrimination in human decisions. There is however an issue in making this parallel---we explain it below on a police search for contraband example. 

Discrimination in the US Law is based on two main concepts, namely \emph{disparate treatment} and \emph{disparate impact}. Disparate impact refers to an apparently neutral policy that adversely affects a protected group more than another group. Evidence that a human decision making has an unjustified disparate impact is often provided using the Outcome Test, which consists in comparing the success rate of decisions. For example, in a police search for contraband scenario, the Outcome Test would check whether minorities ($A=1$) who are searched ($\bar Y=1$) are found to possess contraband ($Y=1$) at the same rate as whites ($A=0$) who are searched, i.e. whether $\mathbb{P}(Y=1|\bar Y=1,A=0)=\mathbb{P}(Y=1|\bar Y=1,A=1)$.
Let $C$ denote the set of characteristics inducing probability of possessing contraband $r^n=\mathbb{P}(Y=1|C=c^n)$. A finding that searches for minorities are systematically less productive than searches for whites can be considered as evidence that the police applies different thresholds $\tau$ when searching, i.e. that $\bar y^n=\1_{r^n>\tau_1}$ if $a^n=1$ whilst $\bar y^n=\1_{r^n>\tau_0}$ if $a^n=0$ with $\tau_1<\tau_0$.

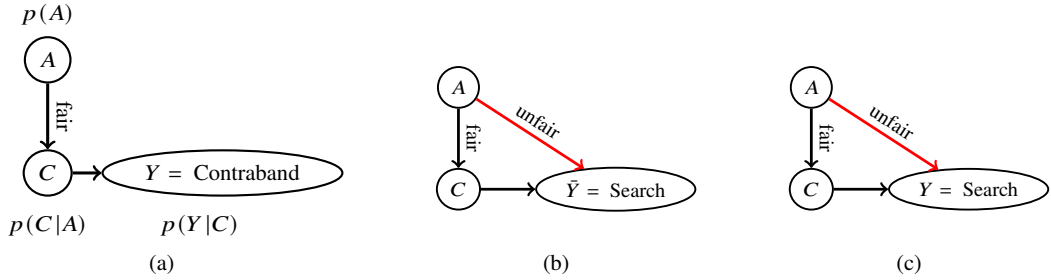
\begin{figure}[t]
\centering
\subfigure[]{
\scalebox{1}{
\begin{tikzpicture}[dgraph]
\node[ocont] (A) at (0,1.5) {$A$};
\node[ocont] (C) at (0,0) {$C$};
\node[ocont] (Y) at (2.3,0) {$Y=\textrm{ Contraband}$};
\node[] at (0,2.1) {$p(A)$};
\node[] at (0,-0.7) {$p(C|A)$};
\node[] at (2,-0.7) {$p(Y|C)$};
\draw[line width=1.15pt](A)--node[sloped,above,black]{fair}++(C);
\draw[line width=1.15pt](C)--(Y);
\end{tikzpicture}}}
\hskip1cm
\subfigure[]{
\scalebox{0.9}{
\begin{tikzpicture}[dgraph]
\node[ocont] (A) at (0,1.5) {$A$};
\node[ocont] (C) at (0,0) {$C$};
\node[ocont] (Y) at (2.3,0) {$\bar Y=\textrm{ Search}$};
\node[] at (0,2.1) {};
\node[] at (0,-0.7) {};
\node[] at (2,-0.7) {};
\draw[line width=1.15pt](A)--node[sloped,above,black]{fair}++(C);
\draw[line width=1.15pt](C)--(Y);
\draw[line width=1.15pt,red](A)--node[sloped,above,black]{unfair}++(Y);
\end{tikzpicture}}}
\hskip1cm
\subfigure[]{
\scalebox{0.9}{
\begin{tikzpicture}[dgraph]
\node[ocont] (A) at (0,1.5) {$A$};
\node[ocont] (C) at (0,0) {$C$};
\node[ocont] (Y) at (2.3,0) {$Y=\textrm{ Search}$};
\node[] at (0,2.1) {};
\node[] at (0,-0.7) {};
\node[] at (2,-0.7) {};
\draw[line width=1.15pt](A)--node[sloped,above,black]{fair}++(C);
\draw[line width=1.15pt](C)--(Y);
\draw[line width=1.15pt,red](A)--node[sloped,above,black]{unfair}++(Y);
\end{tikzpicture}}}
\caption{CBNs describing a police search for contraband example.}
\label{fig:PoliceSearch}
\end{figure}

This scenario can be represented using the causal Bayesian network in Fig. \ref{fig:PoliceSearch}a. A Bayesian network is a \emph{directed acyclic graph} where nodes and edges represent random variables and statistical dependencies respectively. Each node $X_i$ in the graph is associated with the conditional distribution $p(X_i|\textrm{pa}(X_i))$, where $\textrm{pa}(X_i)$ is the set of \emph{parents} of $X_i$.
The joint distribution of all nodes, $p(X_1, \ldots, X_I)$, is given by the product of all conditional distributions, i.e.~$p(X_1,\ldots,X_I)=\prod_{i=1}^Ip(X_i|\textrm{pa}(X_i))$.
A \emph{path} from $X_i$ to $X_j$ is a sequence of linked nodes starting at $X_i$ and ending at $X_j$. A path is called \emph{directed} if the links point from preceding towards following nodes in the sequence. A node $X_i$ is an \emph{ancestor} of a node $X_j$ if there exists a directed path from $X_i$ to $X_j$. In this case, $X_j$ is a \emph{descendant} of $X_i$.
When equipped with causal semantic, namely when representing the data-generation mechanism, Bayesian networks can be used to visually express causal relationships.
More specifically, causal Bayesian networks enable us to give a graphical definition of causes: If there exists a directed path from $A$ to $Y$, then $A$ is a potential cause of $Y$. In CBNs, directed paths are called \emph{causal} paths.

The CBN of Fig. \ref{fig:PoliceSearch}a has joint distribution $p(A,C,Y)=p(Y|C)p(C|A)p(A)$. $A$ is a potential cause of $Y$, because the path $A\rightarrow C\rightarrow Y$ is causal. 
The influence of $A$ on $Y$ through $C$ is considered legitimate as indicated by the label 'fair'. This in turn means that the dependence of $R$ on $A$ is considered fair. We can interpret the Outcome Test in the CBN framework as an indirect way to understand whether the police is introducing an unfair causal path $A\rightarrow \bar Y$ as depicted in Fig. \ref{fig:PoliceSearch}b, namely whether it bases the decision on whether to search a person on race $A$, in addition to $C$ (i.e. if two individuals with the same characteristics are treated differently if of different race).

Predictive Parity is often seen as the equivalent of the Outcome Test for ML classifiers. However, the problem in making this parallel is that, in the Outcome Test, $Y$ corresponds to actual possession of contraband whilst, in the dataset used to train a classifier, $Y$ could instead correspond to police search and the police could be discriminating by using different thresholds for searching minorities and whites. This scenario is depicted in Fig. \ref{fig:PoliceSearch}c. In this scenario in which the training data contains an unfair path $A\rightarrow Y$, Predictive Parity is not a meaningful fairness goal.

More generally, in the case in which at least one causal path from $A$ to $Y$ is unfair, both EFPRs/EFNRs and Predictive Parity are inappropriate criteria, as they do not require the unfair influence of $A$ on $Y$ to be absent from the prediction $\hat Y$ (e.g.~a perfect model ($\hat Y = Y$) would automatically satisfy EFPRs/EFNRs and Predictive Parity, but would contain the unfair influence). 

This observation is particularly relevant to the recent debate surrounding the Correctional Offender Management Profiling for Alternative Sanctions (COMPAS) pretrial risk assessment tool developed by Equivant (formerly Northpointe) and deployed in Broward County in Florida.
The debate was triggered by an expos\'e from investigative journalists at ProPublica \cite{angwin2016machine}. ProPublica claimed that COMPAS did not satisfy EFPRs and EFNRs, as $\textrm{FPRs }=44.9\%$ and $\textrm{FNRs }=28.0\%$ for black defendants, whilst $\textrm{FPRs }=23.5\%$ and $\textrm{FNRs }=47.7\%$ for white defendants. 
This evidence led ProPublica to conclude that COMPAS had a disparate impact on black defendants, leading to public outcry over potential biases in risk assessment tools and machine learning writ large.
In response, Equivant published a technical report \cite{dieterich2016compas} refuting the claims of bias made by ProPublica and concluded that COMPAS is sufficiently calibrated, in the sense that it satisfies Predictive Parity at key thresholds. As previous research has shown \cite{isaac2017hope,lum2016predict,selbst2017disparate}, modern policing tactics center around targeting a small number of neighborhoods---often disproportionately populated by non-white and low income residents---with recurring patrols and stops. This uneven distribution of police attention, as well as other factors such as funding for pretrial services \cite{koepke2017danger,stevenson2017assessing}, can be rephrased in the language of CBNs as indicating the presence of a direct path $A\rightarrow Y$ (through unobserved neighborhood) in the CBN~representing the data-generation mechanism, as well as an influence of $A$ on $Y$ through the set of variables containing number of prior arrests that are used to form a prediction $\hat Y$ of $Y$. The fairness of these paths is questionable, indicating that EFPRs/EFNRs and Predictive Parity are inappropriate criteria. More generally, these observations indicate that the fairness debate surrounding COMPAS gave insufficient consideration to the patterns of unfairness underlying the training data.\\

The characterization of unfairness in a dataset as the presence of unfair causal paths has enabled us to demonstrate the danger in making parallels between tests for discrimination in ML and human decision making, and that a correct use of fairness definitions concerned with statistical properties of $\hat Y$ with respect to $A$ requires an understanding of the patterns of unfairness underlying the training data. Deciding whether a path is fair or unfair requires careful ethical and sociological considerations and/or might not be possible from a dataset alone. Furthermore, a path could also be only partially fair, a case that we omitted here for simplicity. Despite these limitations, this viewpoint enables us to use CBNs as simple and intuitive visual tool to reason about fairness. 
\subsection{Causal Bayesian Networks: A Quantitative Tool for (Un)fairness}
\label{sec:CBNs:QT}
In this section we discuss how CBNs enable us to quantify unfairness in complex scenarios and to develop techniques for addressing it. More details can be found in \cite{chiappa2019path,chiappa2019causal,chiappa2020general}. 
\subsubsection{Background on Causal Bayesian Networks\label{sec:BCBNs}} 
\noindent
\begin{wrapfigure}[5]{l}{0.2\textwidth}
\vspace{-0.6cm}
\centering
\scalebox{0.78}{
\begin{tikzpicture}[dgraph]
\node[ocont] (C) at (1,1.5) {$C$};
\node[ocont] (A) at (0,0) {$A$};
\node[ocont] (Y) at (2,0) {$Y$};
\draw[line width=1.15pt](C)--(A);\draw[line width=1.15pt](C)--(Y);\draw[line width=1.15pt](A)--(Y);
\end{tikzpicture}}
\label{fig:CE}
\end{wrapfigure}
\noindent {\bf Causal Effect.} Consider the CBN on the left, which contains one causal and one non-causal path from $A$ to $Y$, given by $A\rightarrow Y$ and $A\leftarrow C\rightarrow Y$ respectively. Whilst the conditional distribution $p(Y|A)$ measures information from $A$ to $Y$ traveling through both paths\footnote{This is the case as the non-causal path $A\leftarrow C\rightarrow Y$ is \emph{open}.}, the
\emph{causal effect} of $A$ on $Y$, denoted with $p_{\rightarrow A}(Y|A)$, measures information traveling through the causal path $A\rightarrow Y$ only. Thus, the causal effect of $A$ on $Y$ can be seen as the conditional distribution of $Y$ given $A$ restricted to causal paths.\\

\noindent {\bf Path-Specific Effect.} Let us define with $Y_{\rightarrow A=a}$ the random variable with distribution $p(Y_{\rightarrow A=a}) = p_{\rightarrow A=a}(Y|A=a)$.
$Y_{\rightarrow A=a}$ is called \emph{potential outcome} and we refer to it with the shorthand $Y_{\rightarrow a}$.\\[-10pt]

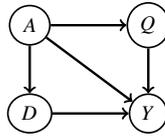
\begin{wrapfigure}[5]{l}{0.2\textwidth}
\vspace{-0.6cm}
\centering
\scalebox{0.78}{
\begin{tikzpicture}[dgraph]
\node[ocont] (A) at (0,1.5) {$A$};
\node[ocont] (Q) at (2,1.5) {$Q$};
\node[ocont] (D) at (0,0) {$D$};
\node[ocont] (Y) at (2,0) {$Y$};
\draw[line width=1.15pt](A)--(D);
\draw[line width=1.15pt](A)--(Y);
\draw[line width=1.15pt](D)--(Y);
\draw[line width=1.15pt](Q)--(Y);
\draw[line width=1.15pt](A)--(Q);
\end{tikzpicture}}
\end{wrapfigure}

\noindent Potential outcomes can be extended to allow to separate the causal effect along different causal paths. Consider a college admission scenario described by the CBN on the left, where $A$ indicates gender, $Q$ qualifications, $D$ choice of department, and $Y$ admission. The causal path $A\rightarrow Y$ represents direct influence of gender $A$ on admission $Y$, capturing the fact that two individuals with the same qualifications and applying to the same department can be treated differently depending on their gender. The indirect causal path $A\rightarrow Q \rightarrow Y$ represents influence of $A$ on $Y$ through $Q$, capturing the fact that female applicants might have different qualifications than male applicants. The indirect causal path $A\rightarrow D \rightarrow Y$ represents influence of $A$ on $Y$ through $D$, capturing the fact that female applicants more often apply to certain departments. 
Let us indicate with $A=a$ and $A=\bar a$ female and male applicants respectively. The \emph{path-specific potential outcome} $Y_{\rightarrow \bar a}(Q_{\rightarrow a},D_{\rightarrow a})$ is defined as the random variable with distribution equal to the 
conditional distribution of $Y$ given $A$, restricted to causal paths, for which $A$ has been set to the value $\bar a$ along $A\rightarrow Y$, and to the value $a$ along $A\rightarrow Q \rightarrow Y$ and $A\rightarrow D \rightarrow Y$. This distribution is given by $p(Y_{\rightarrow \bar a}(Q_{\rightarrow a},D_{\rightarrow a}))=\sum_{Q,D}p(Y|A=\bar a,Q,D)p(Q|A=a)p(D|A=a)$.\\[-5pt]

The \emph{Path-Specific Effect} (PSE) of $A=\bar a$ with respect to $A=a$, defined as
\begin{align}
\textrm{PSE}_{a\bar a} = \mathbb{E}_{p(Y_{\rightarrow\bar a}(Q_{\rightarrow a},D_{\rightarrow a}))}[Y_{\rightarrow\bar a}(Q_{\rightarrow a},D_{\rightarrow a})]-\mathbb{E}_{p(Y_{\rightarrow a})}[Y_{\rightarrow a}]
\label{eq:PSE}
\end{align} 
where $\mathbb{E}_{p(X)}[X]$ denotes expectation, is used to give an estimate of the average effect of $A$ on $Y$ along $A\rightarrow Y$. \\

\noindent {\bf Path-Specific Counterfactual Distribution.}
By conditioning $Y_{\rightarrow \bar a}(Q_{\rightarrow a},D_{\rightarrow a})$ on information from a specific individual, e.g. from a female individual $\{a^n=a, q^n, d^n, y^n\}$ who was not admitted, we can answer the counterfactual question of whether that individual would have been admitted had she been male along the path $A\rightarrow Y$. \\[-10pt]

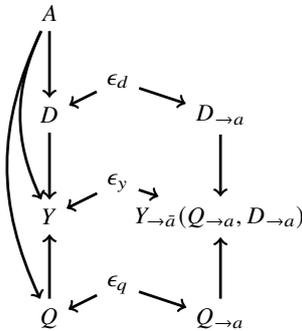
\begin{wrapfigure}[9]{l}{0.3\textwidth}
\vspace{-0.5cm}
\centering
\scalebox{0.9}{
\begin{tikzpicture}[dgraph]
\node[] (eD) at (1,2) {$\epsilon_d$};
\node[] (eY) at (1,0.5) {$\epsilon_y$};
\node[] (eQ) at (1,-1) {$\epsilon_q$};
\node[] (A) at (0,3) {$A$};
\node[] (D) at (0,1.5) {$D$};
\node[] (Y) at (0,0) {$Y$};
\node[] (Q) at (0,-1.5) {$Q$};
\node[] (D*) at (2.5,1.5) {$D_{\rightarrow a}$};
\node[] (Y*) at (2.5,0) {$Y_{\rightarrow \bar a}(Q_{\rightarrow a},D_{\rightarrow a})$};
\node[] (Q*) at (2.5,-1.5) {$Q_{\rightarrow a}$};
\draw[line width=1.15pt](A)--(D);\draw[line width=1.15pt](A)to [bend left=-25](Y);\draw[line width=1.15pt](D)--(Y);
\draw[line width=1.15pt](A)to [bend left=-25](Q);
\draw[line width=1.15pt](Q)--(Y);
\draw[line width=1.15pt](Q*)--(Y*);\draw[line width=1.15pt](D*)--(Y*);
\draw[line width=1.15pt](eD)--(D);\draw[line width=1.15pt](eD)--(D*);
\draw[line width=1.15pt](eQ)--(Q);\draw[line width=1.15pt](eQ)--(Q*);
\draw[line width=1.15pt](eY)--(Y);\draw[line width=1.15pt](eY)--(Y*);
\end{tikzpicture}}
\end{wrapfigure}

\noindent To understand how the path-specific counterfactual distribution $p(Y_{\rightarrow \bar a}(Q_{\rightarrow a},D_{\rightarrow a})|a^n=a, q^n, d^n, y^n)$ can be computed, consider the following linear structural equation model associated to a CBN with the same structure
\begin{align}
& A\sim\textrm{Bern}(\pi),\nonumber \\
& Q=\theta^q+\theta^q_{a}A+\epsilon_q,\nonumber \\
& D=\theta^d+\theta^d_{a}A+\epsilon_d,\nonumber \\ 
& Y=\theta^y+\theta^y_{a}A+\theta^y_{q}Q+\theta^y_{d}D+\epsilon_y, 
\label{eq:model}
\end{align}
where $\epsilon_q,\epsilon_d$ and $\epsilon_y$ are unobserved independent zero-mean Gaussian variables. The relationships between $A,Q,D,Y$ and $Y_{\rightarrow\bar a}(Q_{\rightarrow a},D_{\rightarrow a})$ in this model can be inferred from the \emph{twin Bayesian network} \cite{pearl2000causality} on the left: In addition to $A,Q,D$ and $Y$, the network contains the variables $Q_{\rightarrow a}$, $D_{\rightarrow a}$ and $Y_{\rightarrow\bar a}(Q_{\rightarrow a},D_{\rightarrow a})$ corresponding to the counterfactual world in which $A=\bar a$ along $A\rightarrow Y$, with
\begin{align}
& Q_{\rightarrow a}=\theta^q+\theta^q_{a}a+\epsilon_q,\nonumber \\ 
& D_{\rightarrow a}=\theta^d+\theta^d_{a}a+\epsilon_d,\nonumber\\
& Y_{\rightarrow \bar a}(Q_{\rightarrow a},D_{\rightarrow a})=\theta^y+\theta^y_a\bar a+\theta^y_{q}Q_{\rightarrow a}+\theta^y_{d}D_{\rightarrow a} +\epsilon_y. 
\end{align}

\noindent The two groups of variables $A,Q,D,Y$ and $Q_{\rightarrow a},D_{\rightarrow a},Y_{\rightarrow \bar a}(Q_{\rightarrow a},D_{\rightarrow a})$ are connected by $\epsilon_d, \epsilon_q$ and $\epsilon_y$, indicating that the factual and counterfactual worlds share the same unobserved randomness. From the twin network we can deduce that $Y_{\rightarrow \bar a}(Q_{\rightarrow a},D_{\rightarrow a})~\independent~\{A,Q,D,Y\}~|~\epsilon=\{\epsilon_q,\epsilon_d, \epsilon_y\}$,
and therefore that $p(Y_{\rightarrow \bar a}(Q_{\rightarrow a},D_{\rightarrow a})|A=a, Q=q^n, D=d^n, Y=y^n)$ can be expressed as
\begin{align}
\int_{\epsilon} p(Y_{\rightarrow \bar a}(Q_{\rightarrow a},D_{\rightarrow a})| \epsilon, \cancel{a}, \cancel{q^n}, \cancel{d^n}, \cancel{y^n})p(\epsilon|a, q^n, d^n, y^n) .
\label{eq:counterf}
\end{align}
As $p(\epsilon|a, q^n, d^n, y^n)$ factorizes over $\epsilon_q,\epsilon_d,\epsilon_y$, $p(\epsilon^n_q|a, q^n, d^n, y^n)=\delta_{\epsilon^n_q=q^n-\theta^q-\theta^q_aa}$, $p(\epsilon^n_d|a, q^n, d^n, y^n)=\delta_{\epsilon^n_d=d^n-\theta^d-\theta^d_aa}$, and $p(\epsilon^n_y|a, q^n, d^n, y^n)=\delta_{\epsilon^n_y=y^n-\theta^y-\theta^y_aa-\theta^y_qq^n-\theta^y_dd^n}$, we obtain 
\begin{align}
p(Y_{\rightarrow \bar a}(Q_{\rightarrow a},D_{\rightarrow a})|a, q^n, d^n, y^n)=\delta_{Y_{\rightarrow \bar a}(Q_{\rightarrow a},D_{\rightarrow a})=y^n+\theta^y_a(\bar a-a)}.
\end{align}
Indeed, by expressing $Y_{\rightarrow \bar a}(Q_{\rightarrow a},D_{\rightarrow a})$ as a function of $\epsilon^n_q, \epsilon^n_d$ and $\epsilon^n_y$, we obtain
\begin{align}
\theta^y+\theta^y_a\bar a+\theta^y_{q}(\theta^q+\theta^q_{a}a+\epsilon^n_q)+\theta^y_{d}(\theta^d+\theta^d_{a}a+\epsilon^n_d)+\epsilon^n_y&=\theta^y+\theta^y_a\bar a+\theta^y_{q}q^n+\theta^y_{d}d^n+\epsilon^n_y\nonumber\\
&=y^n+\theta^y_a(\bar a-a)\nonumber\\
&:=y^n_{\rightarrow \bar a}.
\end{align}
Therefore, the path-specific counterfactual distribution concentrates its mass on a single value $y^n_{\rightarrow \bar a}$ given by correcting $y^n$ through replacing $\theta^y_aa$ with $\theta^y_a\bar a$. Notice that $\theta^y_a(\bar a-a)=\textrm{PSE}_{a\bar a}$. With the notation $Y=f_{\theta^y}(A,Q,D,\epsilon_y)$, we have $y^n_{\rightarrow\bar a}=f_{\theta^y}(\bar a,q^n,d^n,\epsilon^n_y)$, which highlights that the desired corrected version of $y^n$ is obtained by using $\bar a$, $q^n,d^n$ and $\epsilon^n_y$ in the structural equation of $Y$.

In the more complex case in which we are interested in $p(Y_{\rightarrow \bar a}(Q_{\rightarrow a},D_{\rightarrow \bar a})|A=a, Q=q^n, D=d^n, Y=y^n)$, a similar reasoning would give
\begin{align}
& \theta^y+\theta^y_a\bar a+\theta^y_{q}(\theta^q+\theta^q_{a}a+\epsilon^n_q)+\theta^y_{d}(\theta^d+\theta^d_{a}\bar a+\epsilon^n_d)+\epsilon^n_y =
\theta^y+\theta^y_a\bar a+\theta^y_{q}q^n+\theta^y_{d}(\theta^d+\theta^d_{a}\bar a+\epsilon^n_d)+\epsilon^n_y \nonumber \\
&=y^n+(\theta^y_a+\theta^y_{d}\theta^d_{a})(\bar a-a)\nonumber\\
&:=y^n_{\rightarrow\bar a}(Q_{\rightarrow a},D_{\rightarrow \bar a}).
\label{eq:counter}
\end{align}
Notice that $(\theta^y_a+\theta^y_{d}\theta^d_{a})(\bar a-a)=\textrm{PSE}_{a\bar a}$. With the notation $d^n_{\rightarrow \bar a}=f_{\theta^d}(\bar a,\epsilon^n_d)$, we have $y^n_{\rightarrow\bar a}(Q_{\rightarrow a},D_{\rightarrow \bar a})=f_{\theta^y}(\bar a,q^n,d^n_{\rightarrow\bar a},\epsilon^n_y)$, which highlights that the corrected version of $y^n$ is obtained by using $\bar a$, $q^n,\epsilon^n_y$, and a corrected version of $d^n$ in the structural equation of $Y$. More generally, the structural equation of $Y$ requires corrections for all variables that are descendants of $A$ along unfair paths---such corrections being obtained with similar modifications of the corresponding structural equations.

In a more complex scenario in which e.g. $D=f_{\theta^d}(A, \epsilon_d)$ for a non-invertible non-linear function $f_{\theta^d}$, we can sample $\epsilon^{n,m}_d$ from $p(\epsilon_d|a, q^n, d^n, y^n)$ and perform a Monte-Carlo approximation of Eq.~\eqref{eq:counterf}, obtaining
\begin{align}
y^n_{\rightarrow\bar a}(Q_{\rightarrow a},D_{\rightarrow \bar a})=
\frac{1}{M}\sum_{m=1}^M f_{\theta^y}(\bar a,q^n,d^{n,m}_{\rightarrow\bar a},\epsilon^{n}_y), \textrm{ where } d^{n,m}_{\rightarrow\bar a}=f_{\theta^d}(\bar a, \epsilon^{n,m}_d).
\label{eq:MC}
\end{align}
It is clear from this discussion that, in order to be able to transfer individual-level information from the factual to the counterfactual world for the variables that are descendants of $A$ along unfair paths, an explicit representation of the unobserved randomness as done in structural equation models is required.
\subsubsection{Quantify Unfairness in a Dataset}

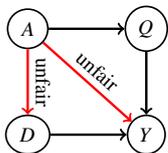
\begin{wrapfigure}[5]{l}{0.2\textwidth}
\vspace{-0.6cm}
\centering
\scalebox{0.78}{
\begin{tikzpicture}[dgraph]
\node[ocont] (A) at (0,1.5) {$A$};
\node[ocont] (Q) at (2,1.5) {$Q$};
\node[ocont] (D) at (0,-0.3) {$D$};
\node[ocont] (Y) at (2,-0.3) {$Y$};
\draw[line width=1.15pt,red](A)--node[sloped,above,black]{unfair}++(D);
\draw[line width=1.15pt,red](A)--node[sloped,above,black]{unfair}++(Y);
\draw[line width=1.15pt](D)--(Y);
\draw[line width=1.15pt](Q)--(Y);
\draw[line width=1.15pt](A)--(Q);
\end{tikzpicture}}
\end{wrapfigure}
%
The path-specific effect and counterfactual distribution introduced above can be used to quantify unfairness in a dataset in complex scenarios, respectively at the population and individual levels.\\[10pt]
{\bf Path-Specific Unfairness.} In the college admission scenario, let us assume that the departmental differences are a result of systemic cultural pressure---i.e. female applicants apply to specific departments at lower rates because of overt or covert societal discouragement---and that for this reason, in addition to the direct path $A\rightarrow Y$, the path $A\rightarrow D$ (and therefore $A\rightarrow D\rightarrow Y$) is deemed unfair. One way to measure unfairness along $A\rightarrow Y$ and $A\rightarrow D\rightarrow Y$ overall population would be to compute the path-specific effect
\begin{align}
\mathbb{E}_{p(Y_{\rightarrow\bar a}(Q_{\rightarrow a},D_{\rightarrow \bar a}))}[Y_{\rightarrow\bar a}(Q_{\rightarrow a},D_{\rightarrow \bar a})]-\mathbb{E}_{p(Y_{\rightarrow a})}[Y_{\rightarrow a}],
\end{align} 
where $p(Y_{\rightarrow\bar a}(Q_{\rightarrow a},D_{\rightarrow \bar a}))=\sum_{Q,D}p(Y|A=\bar a,Q,D)p(Q|A=a)p(D|A=\bar a)$. Notice that, as computing this quantity requires knowledge of the CBN, if the CBN structure is miss-specified or its conditional distributions are poorly estimated, the resulting estimate could be imprecise.\\

\noindent{\bf Path-Specific Counterfactual Unfairness.}
Rather than measuring unfairness along $A\rightarrow Y$ and $A\rightarrow D\rightarrow Y$ overall population, we might want to know whether a specific female applicant $\{a^n=a, q^n, d^n, y^n\}$ who was not admitted would have been admitted had she been male ($A=\bar a$) along the direct path $A\rightarrow Y$ and the indirect path $A\rightarrow D\rightarrow Y$. This question can be answered by estimating the path-specific counterfactual distribution $p(Y_{\rightarrow a}(Q_{\rightarrow a},D_{\rightarrow \bar a})|A=a, Q=q^n, D=d^n, Y=y^n)$ (e.g. $y^n_{\rightarrow\bar a}(Q_{\rightarrow a},D_{\rightarrow \bar a})$ as in Eq. (\ref{eq:MC})). 
Notice that the outcome in the actual world, $y^n$, corresponds to $p(Y_{\rightarrow a}(Q_{\rightarrow a},D_{\rightarrow a})|A=a, Q=q^n, D=d^n, Y=y^n)$.

\subsubsection{Imposing Fairness in a Model}
In addition to quantify unfairness in a dataset, path-specific effects and counterfactual distributions can also be used to impose fairness in a ML model.

An ML model learned using a dataset from the college admission scenario would often output an estimate of the probability that individual $n$ belongs to class 1, $s^n = \bar p(Y=1|A=a^n,Q=q^n,D=d^n)$, where $\bar p$ indicates the estimate of $p$. We denote with $s^n_{\rightarrow \bar a}(Q_{\rightarrow a},D_{\rightarrow \bar a})$ the model estimated probability that a female applicant $\{a^n=a,q^n,d^n\}$ would have been admitted in a counterfactual world in which she were male along $A\rightarrow Y$ and $A\rightarrow D \rightarrow Y$, i.e.
\begin{align}
s^n_{\rightarrow \bar a}(Q_{\rightarrow a},D_{\rightarrow \bar a}) =\bar p(Y_{\rightarrow \bar a}(Q_{\rightarrow a},D_{\rightarrow \bar a})=1|A=a^n,Q=q^n,D=d^n),
\end{align}
and with $S_{\rightarrow\bar a}(Q_{\rightarrow a},D_{\rightarrow \bar a})$ the corresponding random variable. Notice that, unlike above, we do not condition on $y^n$.\\

\noindent {\bf Path-Specific Fairness.}
We can use the path-specific effect to formalize the requirement that the influence of $A$ along the unfair causal paths $A\rightarrow Y$ and $A\rightarrow D \rightarrow Y$ should be absent from the model, by requiring that the path-specific effect 
\begin{align}
\mathbb{E}_{\bar p(Y_{\rightarrow\bar a}(Q_{\rightarrow a},D_{\rightarrow \bar a}))}[Y_{\rightarrow\bar a}(Q_{\rightarrow a},D_{\rightarrow \bar a})]-\mathbb{E}_{\bar p(Y_{\rightarrow a})}[Y_{\rightarrow a}],
\end{align}
should be zero. This criterion was called
\emph{Path-Specific Fairness} (PSF) in \cite{chiappa2020general}. The work in \cite{nabi2018fair}
introduces a method to achieve PSF based on enforcing the path-specific effect to be small during model training. When deploying the model, this method requires integrating out all variables that are descendants of $A$ along unfair causal paths (in this case $D$) to correct for the unfairness in their realizations, with consequent potential loss in accuracy.\\

\noindent {\bf Path-Specific Counterfactual Fairness.}
To overcome this issue, the work in \cite{chiappa2019path} proposes to instead correct the model output $s^n$ into its path-specific counterfactual $s^n_{\rightarrow \bar a}(Q_{\rightarrow a}, D_{\rightarrow \bar a})$. The resulting model is said to satisfy  \emph{Path-Specific Counterfactual Fairness}. 
%
%
Following a similar reasoning as in Sect. \ref{sec:BCBNs}, we obtain
\begin{align*}
s^n_{\rightarrow \bar a}(Q_{\rightarrow a}, D_{\rightarrow \bar a})\approx \frac{1}{M}\sum_{m=1}^M \bar p(Y=1| A=\bar a, Q=q^n, D=d^{n,m}_{\rightarrow \bar a}),
\end{align*}
where $d^{n,m}_{\rightarrow\bar a}=\bar f_{\theta^d}(\bar a, \epsilon^{n,m}_d)$ (or $\sim \bar p(D|\bar a, \epsilon^{n,m}_d)$). 
Therefore, the desired path-specific counterfactual is obtained by conditioning $Y$ on the non-descendants of $A$, on the descendants of $A$ that are only fairly influenced by $A$, and on corrected versions of the descendants of $A$ that are unfairly influenced by $A$. 
\section{Methods for Imposing Fairness in a Model}
\label{sec:Methods}
From a procedural viewpoint, methods for imposing fairness can roughly be grouped into pre-processing, in-processing, and post-processing methods. Pre-Processing methods consist in transforming the training data to remove undesired biases~\cite{beutel2017data,calders2009building,calmon2017optimized,chierichetti2017fair,edwards2016censoring,feldman2015computational,feldman2015certifying,fish2015fair,gordaliza2019obtaining,hajian2012methodology,hajian2014generalization,hajian2011rule,hajian2015discrimination,johndrow2019algorithm,kamiran2009classifying,kamiran2010classification,kamiran2012data,louizos2016variational,lum2016statistical,luong2011k,madras2018learning,mancuhan2012discriminatory,mancuhan2014combating,mcnamara2019costs,mcnamara2017provably,song2018learning,wang2018invariant,zemel2013learning,zhang2017achieving,zhang2017causal,vzliobaite2011handling}. The resulting transformations can then be used to train a ML model in a standard way. 
In-Processing methods enforce a model to produce fair outputs through imposing fairness constraints into the learning mechanism. Some methods transform the constrained optimization problem via the method of Lagrange multipliers~\cite{agarwal2018reductions,berk2017convex,corbett-davies2017algorithmic,cotter2018training,cotter2019two,goh2016satisfying,kearns2018preventing,narasimhan2018learning,zafar2017fairness,zafar2019fairness} or add penalties to the objective~\cite{alabi2018unleashing,bechavod2018Penalizing,calders2013controlling,donini2018empirical,dwork2018decoupled,fukuchi2015prediction,gillen2018online,grgic2017fairness,hebert2017calibration,heidari2018fairness,heidari2018moral,hu2019fair,jabbari2017fairness,johnson2016impartial,joseph2016rawlsian,joseph2016fairness,kamiran2012decision,kamishima2012fairness,kamishima2013independence,kilbertus2017avoiding,komiyama2018nonconvex,luo2015discrimination,madras2018predict,mary2019fairness,nabi2019learning,nabi2018fair,narasimhan2018learning,olfat2017spectral,oneto2019taking,quadrianto2017recycling,speicher2018unified,williamson2019fairness,yang2017measuring,yao2017beyond,yona2018probably}, others use adversary techniques to maximize the system ability to predict the target while minimizing the ability to predict the sensitive attribute~\cite{zhang2018mitigating}. Post-Processing methods consist in transforming the model outputs in order to make them fair~\cite{adler2018auditing,ali2019loss,chzhen2019leveraging,doherty2012information,feldman2015computational,fish2016confidence,hajian2012injecting,hardt2016equality,kim2018fairness,kim2019multiaccuracy,kusner2017counterfactual,noriega2019active,pedreschi2009measuring,pleiss2017fairness,raff2018fair,wu2016using}.

Notice that this grouping is imprecise and non-exhaustive. For example, the method in \cite{chiappa2019path} imposes constraints into the learning mechanism during training in order to ensure that requirements for correct post-processing are met. In addition, whilst not tested for this purpose, the methods to achieve (Path-Specific) Counterfactual Fairness in \cite{chiappa2019path,kusner2017counterfactual} generate corrected model outputs for fairness through correcting the variables used to generate those outputs. As such, they could be used to generate fair datasets, and therefore be categorized as pre-processing methods.

From a setting viewpoint, methods for imposing fairness have for long almost entirely focused on binary classification with categorical sensitive attributes (see \cite{donini2018empirical} for a review of this setting). More recently, regression \cite{berk2017convex,calders2013controlling,fitzsimons2018equality,raff2018fair} has also started to be considered, as well as continuous sensitive attributes through discretization~\cite{komiyama2017two,komiyama2018nonconvex,perez-suay2017fair,yona2018probably}. Most methods are still fragmented and compartmentalized in terms of task and sensitive attribute types, as well as of fairness definitions considered, and most lack consistency results. These limits are partially addressed in \cite{chiappa2020general,donini2018empirical,komiyama2017two,oneto2019general} by introducing unified frameworks that encompass methods that can deal with different settings and fairness criteria and are accompanied by theoretical guarantees about their fairness properties.

The work in \cite{chiappa2020general} views different fairness desiderata as requiring matching of distributions corresponding to different sensitive attributes, and uses optimal transport theory to achieve that for the binary classification and regression settings, providing one of the few approaches that do not approximate fairness desiderata through requirements on the lower order moments or other functions of distributions corresponding to different sensitive attributes. We present a simple post-processing method derived within this framework in Sect. \ref{sec:CDOT}.

In~\cite{donini2018empirical}, simple notions of fairness are incorporated within the Empirical Risk Minimization framework. This framework is extended to cover the whole supervised learning setting with risk and fairness bounds---implying consistency properties both in terms of fairness measure and risk of the selected model---in~\cite{oneto2019general}. We present this work in Sect. \ref{sec:GFERM}. 

\begin{table}[b]
\centering
\scriptsize
\begin{tabular}{|l|c|c|l|l|l|}
\hline
\rotatebox{90}{Datasets\ } & 
\rotatebox{90}{Reference\ } & 
\rotatebox{90}{Number of Samples\ } & 
\rotatebox{90}{Number of Features\ } & 
\rotatebox{90}{Sensitive Features\ } &
\rotatebox{90}{Task\ } \\
\hline
\hline
\Tstrut xAPI Students Performance & \cite{amrieh2015students} & 480 & 16 & Gender, Nationality, Native-Country & MC \\
NLSY & \cite{bureau2019national} & ${\approx}10$K & & Birth-date, Ethnicity, Gender & BC, MC, R \\
Wine Quality & \cite{cortez2009wine} & 4898 & 13 & Color & MC, R \\
Students Performance & \cite{cortez2014student} & 649 & 33 & Age, Gender & R \\
Drug Consumption & \cite{fehrman2016drug} & 1885 & 32 & Age, Ethnicity, Gender, Country & MC \\
School Effectiveness & \cite{goldstein1987school} & 15362 & 9 & Ethnicity, Gender & R \\
Arrhythmia & \cite{guvenir1998arrhythmia} & 452 & 279 & Age, Gender & MC \\
MovieLens & \cite{harper2016movielens} & 100K & ${\approx}20$ & Age, Gender & R\\
Heritage Health & \cite{heritage2011heritage} & ${\approx}60$K & ${\approx}20$ & Age, Gender & MC, R \\
German Credit & \cite{hofmann1994statlog} & 1K & 20 & Age, Gender/Marital-Stat & MC \\
Student Academics Performance & \cite{hussain2018student} & 300 & 22 & Gender & MC \\
Heart Disease & \cite{janosi1988heart} & 303 & 75 & Age, Gender & MC, R \\
Census/Adult Income & \cite{kohavi1996census} & 48842 & 14 & Age, Ethnicity, Gender, Native-Country & BC \\
COMPAS & \cite{larson2016propublica} & 11758 & 36 & Age, Ethnicity, Gender & BC, MC \\
Contraceptive Method Choice & \cite{lim1997contraceptive} & 1473 & 9 & Age, Religion & MC \\
CelebA Faces & \cite{liu2015celeba} & ${\approx}200$K & 40 & Gender Skin-Paleness, Youth & BC \\
Chicago Faces & \cite{ma2015chicago} & 597 & 5 & Ethnicity, Gender & MC \\
Diversity in Faces & \cite{merler2019diversity} & 1 M & 47 & Age, Gender & MC, R \\
Bank Marketing & \cite{moro2014bank} & 45211 & 17-20 & Age & BC \\
Stop, Question \& Frisk & \cite{new2012stop} & 84868 & ${\approx}100$ & Age, Ethnicity, Gender & BC, MC \\
Communities \& crime & \cite{redmond2009communities} & 1994 & 128 & Ethnicity & R \\
Diabetes US & \cite{strack2014diabetes} & 101768 & 55 & Age, Ethnicity & BC, MC \\
Law School Admission & \cite{wightman1998law} & 21792 & 5 & Ethnicity, Gender & R \\
\Bstrut Credit Card Default & \cite{yeh2016default} & 30K & 24 & Age, Gender & BC \\
\hline
\end{tabular}
\caption{Publicly available datasets commonly used in the ML fairness literature.\label{ref:reviewtable_dataset}}
\end{table}

Most often the goal of current methods for imposing fairness is to create a fair model for a fixed scenario. However, in a large number of real world applications~\cite{OnetoC003,OnetoJ021,OnetoJ015,OnetoC037} using the same model or part of it over different tasks might be desirable. For example, it is common to perform a fine tuning over pre-trained models~\cite{donahue2014decaf}, keeping the internal representation fixed. Unfortunately, fine tuning a model which is fair on a task on novel previously unseen tasks could lead to an unexpected unfairness behaviour (i.e.~discriminatory transfer~\cite{discriminative_transfer} or negative legacy~\cite{kamishima2012fairness}), due to missing generalization guarantees concerning its fairness properties.
To avoid this issue, it is necessary to consider the learning problem in a multitask/lifelong learning framework.
Recent methods leverage task similarity to learn fair representations that provably generalizes well to unseen tasks. Such methods can be seen as a sophisticated form of pre-processing methods. We discuss one such a method in Sect. \ref{sec:LFR}.

Legal requirements often forbid the explicit use of sensitives attributes in the model. Therefore research into developing methods that meet these requirements is important. In Sect. \ref{sec:FairModels:Sens}, we discuss issues with not explicitly using sensitives attributes and introduce a method that does not require the use of them during model deployment.

Before delving into the explanation of specific methods, in Table~\ref{ref:reviewtable_dataset} we give a list of publicly available datasets commonly used in the ML fairness literature.
\subsection{Constraints on Distributions with Optimal Transport}
\label{sec:CDOT}
Most methods to obtain fair models impose approximations of fairness desiderata through constraints on lower order moments or other functions of distributions corresponding to different sensitive attributes (this is also what most popular fairness definitions require). Whilst facilitating model design, not imposing constraints on the full shapes of relevant distributions can be problematic. By matching distributions corresponding to different sensitive attributes either in the space of model outputs or in the space of model inputs (or latent representations of the inputs) using optimal transport theory, the work in \cite{chiappa2020general,jiang2019wasserstein} introduces an approach to fair classification and regression that is applicable to many fairness criteria. In this section, we describe a simple post-processing method to achieve Strong Demographic Parity that was derived within this work.\\

\noindent {\bf Strong Demographic Parity.} Let us extend the notation of Sect. \ref{sec:CBNs:VT} to include regression and multiple, possibly non-binary, sensitive attributes. That is, let us assume that, in the dataset $\mathcal{D} =\{(a^n,x^n,y^n)\}_{n=1}^N$, $y^n$ can be continuous or categorical, and $a^n\in \mathcal{A}=\mathbb{N}^k$ (where element $a^n_i$ might correspond e.g. to gender). Regression and classification can be uniquely framed as the task of estimating the probability distribution $p(Y|A,X)$, and by assuming that the model outputs the expectation
\begin{align}
s^n=\mathbb{E}_{\bar p(Y|A=a^n,X=x^n)}[Y]\,,
\end{align}
where $\bar p$ indicates the estimate of $p$ (below we omit the distinction between $p$ and $\bar p$ to simplify the notation). A prediction $\hat y^n$ of $y^n$ is thus obtained as $\hat y^n=s^n$ for the regression case, and as
$\hat y^n=\1_{s^n>\tau}$ for the classification case. 
We denote with $S_{a}$ the output variable corresponding to the group of individuals with sensitive attributes $a$, i.e. with distribution $p(S_a)=p(S|A=a)$ (we also denote this distribution with $p_{S_a}$). We can extend Demographic Parity (Sect. \ref{sec:CBNs:VT}) to this setting by re-phrasing it as the requirement that the expectation of $\hat Y$ should not depend on $A$, i.e.
\begin{align}
\mathbb{E}_{p(\hat Y|A=a)}[\hat Y] = \mathbb{E}_{p(\hat Y|A=\bar a)}[\hat Y], 
\hskip0.1cm \forall a,\bar a\in \mathcal{A}.
\end{align}
In the case of classification, enforcing Demographic Parity at a given threshold $\tau$ does not necessarily imply that the criterion is satisfied for other thresholds. Furthermore, to alleviate difficulties in optimizing on the class prediction $\hat Y$, relaxations are often considered, such as imposing the constraint $\mathbb{E}[S|A=a]=\mathbb{E}[S| A = \bar a]$, $\forall a,\bar a \in \mathcal{A}$ \cite{goh2016satisfying,zafar2017fairnessARXIV}. In the case of regression, whilst Demographic Parity is a commonly used criterion \cite{fitzsimons2018equality}, it represents a limited way to enforce similarity between the conditional distributions $p(S|A)$. 

To deal with these limitations, the work in \cite{chiappa2020general,jiang2019wasserstein} introduced the \emph{Strong Demographic Parity} (SDP) criterion, which requires $S~\independent ~A$, i.e.
\begin{align}
p(S_{a}) = p(S_{\bar a}), \hskip0.1cm \forall a,\bar a\in\mathcal{A},
\end{align}
and an approach to enforce SDP using optimal transport theory \cite{peyre2019computational,villani2009optimal}.\\

\noindent {\bf Optimal Transport.} 
In Monge's formulation \cite{monge1781memoire}, the optimal transport problem consists in transporting a distribution to another one incurring in minimal cost. More specifically, given two distributions $p_{X}$ and $p_{Y}$ on ${\cal X}$ and ${\cal Y}$, the set $\mathcal{T}$ of \emph{transportation maps} from $\mathcal{X}$ to $\mathcal{Y}$ (where each transportation map $T: {\cal X}\rightarrow {\cal Y}$ satisfies $\int_{\mathcal{B}}p_{Y}(y)dy = \int_{T^{-1}(\mathcal{B})}p_{X}(x)dx$ for all measurable subsets $\mathcal{B}\subseteq {\cal Y}$), and a \emph{cost function} $\tc:{\cal X}\times {\cal Y}\rightarrow[0,\infty]$, the optimal transport problem consists in finding the transportation map $T^*$ that minimizes the total transportation cost, i.e. such that
\begin{align}
T^* = \argmin_{T\in\mathcal{T}} \W_\tc(p_{X}, p_{Y} ) = \argmin_{T\in\mathcal{T}} \int\tc(x, T(x)) p_{X}(x) dx.
\end{align}
%
If $\mathcal{X} = \mathcal{Y}$ and $\mathcal{C} = D^p$ for some distance metric $D: \mathcal{X} \times \mathcal{Y} \rightarrow \mathbb{R}$ and $p\geq 1$, then $\W_{\tc}(p_X, p_Y)^{1/p}$ is a valid distance between $p_X$ and $p_Y$. When $\mathcal{X} = \mathcal{Y} = \mathbb{R}^d$ and $\tc( x, y) = \|  x-y \|_{p}^p$, where $\|  \cdot \|_{p}$ indicate the $L^p$ norm, $\W_\tc(p_X, p_Y)$ corresponds to the $p$th power of the Wasserstein-$p$ distance and we adopt the shorthand $\W_p(p_{X}, p_{Y})$ to denote it.\\

\noindent {\bf Post-Processing by Transporting Distributions to their Barycenter.} 
We would like to perform a post-processing of the model outputs to reach SDP by transporting the distribution $p_{S_a}$ of each group output variable $S_a$ to a common distribution $p_{\bar S}$. In order to retain accuracy and induce minimal deviation, we would like to use a transportation map $T^*_a$ such that $T^*_a(S_a)$ remains close to $S_a$.

For regression, $T^*_a$ minimizing $\mathbb{E}_{p_{S_a}}[(S_a-T_a(S_a))^2]$ would satisfy this property in a least-squares sense, giving $T^*_a =\argmin_{T_a\in\mathcal{T}(p_{S_a}, p_{\bar S})}\W_2(p_{S_a}, p_{\bar S})$.
Considering all groups, each weighted by its probability $p_a=p(A=a)$, we obtain that the distribution $p_{\bar S}$ inducing the minimal deviation from $S$ is given by 
\begin{align}
    p_{\bar S} &= \argmin_{p^*} \sum_{a\in\mathcal{A}} p_{a} \W_2(p_{S_a}, p^*).
\end{align}
This distribution coincides with the Wasserstein-2 barycenter with weights $p_{a}$. 

For classification, using instead the $L^1$ norm gives the Wasserstein-1 barycenter. This has the desirable property of inducing the minimal number of class prediction changes in expectation. 
Indeed, a class prediction $\hat y=\mathbbm{1}_{s_a>\tau}$ changes due to transportation $T(s_a)$ if and only if $\tau\in\big(m_{s_a}^T, M_{s_a}^T\big)$ where $m_{s_a}^T = \min[s_a, T(s_a)]$ and $M_{s_a}^T = \max [s_a, T(s_a)]$. This observation leads to the following result.
\begin{proposition}
\label{4.3:prop:1}
Let $S_a$ and $\bar S$ be two output variables with values in $[0,1]$ and with distributions $p_{S_a}$ and $p_{\bar S}$,
and let $T: [0,1] \rightarrow [0,1]$ be a transportation map satisfying $\int_{\mathcal{B}}p_{\bar S}(y)dy = \int_{T^{-1}(\mathcal{B})}p_{S_a}(x)dx$ for any measurable subset $\mathcal{B}\subset {[0,1]}$.
The following two quantities are equal:
\begin{enumerate}
\item $\mathcal{W}_1(p_{S_a}, p_{\bar S}) = \min\limits_{T} \int_{x\in[0,1]} |x-T(x)|p_{S_a}(x)dx$,
\item Expected class prediction changes due to transporting $p_{S_a}$ into $p_{\bar S}$ through the map $T^*$, $$\mathbb{E}_{\tau\sim U([0,1]), x\sim p_{S_a}} \mathbb{P}\big(\tau\in\big(m_x^{T^*}, M_x^{T^*}\big)\big),$$
\end{enumerate}
\end{proposition}
where $U$ indicate the uniform distribution. The proof is reported in~\cite{jiang2019wasserstein}.

In summary, the optimal post-processing procedure to ensure fairness whilst incurring in minimal model deviation is to transport all group distributions $p_{S_a}$ to their weighted barycenter distribution $p_{\bar S}$.\\

\noindent {\bf Partial Transportation for Fairness-Accuracy Trade-Off.}
Whilst the approach described above allows to achieve SDP by retaining as much accuracy as possible, in some cases we might want to trade-off a certain amount of fairness for higher accuracy. In the remainder of this section, we explain how to obtain an optimal trade-off for the case of the Wasserstein-2 metric space. 

Not achieving SDP implies that each $p_{S_a}$ is transported to a distribution $p_{S_a^*}$ that does not match the barycenter $p_{\bar S}$. A valid measure of deviation from SDP is  $d_{\textrm{pair}} = \sum_{a \neq \bar a} \W_2(p_{S_a^*}, p_{S_{\bar a}^*})$ since $d_{\textrm{pair}} = 0 \Leftrightarrow p_{S_a^*} = p_{S_{\bar a}^*}, \forall a,\bar a\in\mathcal{A}$. For any distribution $p$,
by the triangle and Young's inequalities, 
\begin{align}
d_{\textrm{pair}} &\leq \sum_{a \neq \bar a} \Big(\sqrt{\W_2(p_{S_a^*}, p)} +  \sqrt{\W_2(p, p_{S^*_{\bar a}})}\Big)^2 \nonumber\\
&\leq \sum_{a \neq \bar a} 2\Big(\W_2(p_{S_a^*}, p) + \W_2(p, p_{S^*_{\bar a}})\Big)
= 4(|\mathcal{A}|-1) \sum_{a\in\mathcal{A}}\W_2(p_{S_a^*}, p)\,.
\end{align}
By the definition of the barycenter, this upper bound reaches its minimum when $p=p_{\bar S}$. We call this tightest upper bound pseudo-$d_{\textrm{pair}}$ and use it to derive optimal trade-off solutions. 
 
For any $r\in \mathbb{R}_+$, we say that pseudo-$d_{\textrm{pair}}$ satisfies the \emph{$r$-fairness} constraint when it is smaller than $r$.
To reach optimal trade-offs, we are interested in transporting $p_{S_a}$ to $p_{S_a^*}$ under the {\it $r$-fairness} constraint while minimizing the deviation from $S$, $\min_{p_{S_a}^*}\sum_{a\in\mathcal{A}} p_a \W_2(p_{S_a}, p_{S_a^*})$. Assuming disjoint groups, we can optimize each group transportation in turn independently. The $r$-fairness constraint on a single group $a$ becomes $\W_2(p_{S_a^*}, p_{\bar S}) \leq r' - d'$, where $r' = r/(4|\mathcal{A}|-4)$ and $d' = \sum_{\bar a\in \mathcal{A}\setminus \{a\}} \W_2(p_{S^*_{\bar a}}, p_{\bar S})$. 
Satisfying this constraint corresponds to transporting $p_{S_a}$ to the ball with center $p_{\bar S}$ and radius $r' - d'$ in the Wasserstein-2 metric space. To achieve the optimal trade-off, we need to transport $p_{S_a}$ to a destination $p_{S_a^*}$ with minimal $\W_2(p_{S_a}, p_{S_a^*})$.
Thus we want
\begin{align}
\hskip-0.2cm p_{S_a^*} &= \argmin_{p^*~\text{s.t.}~\W_2(p^*, p_{\bar S}) \leq r' - d'} \hskip-0.2cm p_a\W_2(p_{S_a}, p^*)= \argmin_{p^*~\text{s.t.}~\W_2(p^*, p_{\bar S}) \leq r' - d'} \hskip-0.3cm\W_2(p_{S_a}, p^*),
\end{align}
since $p_a$ is constant with respect to $p^*$.
As $\W_2(p_{S_a}, p^*) \geq \big(\sqrt{\W_2(p_{S_a}, p_{\bar S})}-\sqrt{\W_2(p^*, p_{\bar S})}\big)^2$ by triangle inequality, $\W_2(p_{S_a}, p^*)$ reaches its minimum if and only if $p^*$ lies on a shortest path between $p_{S_a}$ and $p_{\bar S}$. Therefore it is optimal to transport $p_{S_a}$ along any shortest path between itself and $p_{\bar S}$ in the Wasserstein-2 metric space.
\subsubsection{Wasserstein-2 Geodesic Method} 
In the univariate case, we can derive a simple post-processing method for implementing the optimal trade-off between accuracy and SDP described above based on geodesics.

Let the Wasserstein-2 space $\mathcal{P}_2(\mathbb{R})$ be defined as the space of all distributions $p$ on the metric space $\mathbb{R}$ with finite\footnote{This condition is satisfied as we use empirical approximations of distributions.} absolute $p$-th moments, i.e. $\mathbb{E}_{p(s_1)}\lbrack |s_1 - s_0|^p\rbrack < \infty$ for $\forall s_0\in\mathbb{R}$, equipped with the Wasserstein-2  metric. As $\mathbb{R}$ is a geodesic space, i.e. there exists a geodesic between every pair of points in that space, then so is $\mathcal{P}_2(\mathbb{R})$ \cite{lisini2007characterization}. 
\begin{algorithm}[t]
\textbf{Input: Dataset $\mathcal{D} = \{ (a^n, x^n, y^n)\}_{n=1}^N$, number of bins $B$, model outputs $\{s^n\}$, trade-off parameter $t$.} \\
Obtain group datasets $\{\mathcal{D}_a\}$ and barycenter dataset $\bar{\mathcal{D}}$.\newline
Define the $i$-th quantile of $\mathcal{D}_{a}$, as
\begin{align*}
q_{\mathcal{D}_{a}}(i) := \sup \left\{s : \frac{1}{N_{a}}\sum_{n~\text{s.t.}~a^n=a} \1_{s^n \le s} \le \frac{i-1}{B}\right\},
\end{align*}
and its inverse as $q^{-1}_{\mathcal{D}_{a}}(s) := \sup\{ i \in [B] : q_{\mathcal{D}_{a}}(i) \le s\}$. 
\newline
Define $q^{-1}_{\mathcal{D}_{a, t}}(s) := (1-t)q^{-1}_{\mathcal{D}_{a}}(s) + t\, q^{-1}_{\bar{ \mathcal{D}}}(s)$, giving
$q_{\mathcal{D}_{a, t}}(i) = \sup\Big\{ s \in [0,1] : (1-t)q^{-1}_{\mathcal{D}_{a}}(s) + t\, q^{-1}_{\bar{ \mathcal{D}}}(s) \le i\Big\}$. 
\newline
\textbf{Return:} $\Big\{q_{\mathcal{D}_{a, t}}\left( q^{-1}_{\mathcal{D}_{a}}(s^n)\right)\Big\}$.
\caption{Wasserstein-$2$ Geodesic}
\label{Alg:wass2_pp_tradeoff}
\end{algorithm}
Whilst geodesics are only locally shortest paths, shortest paths are always geodesics if they exist.
In the case of $\mathcal{P}_2(\mathbb{R})$, the geodesic between $p_{S_a}$ and $p_{\bar S}$ is unique and can be parametrized by
\begin{align}
P_{S_a, t}^{-1} = (1-t)P_{S_a}^{-1} + t P_{\bar S}^{-1}, \hskip0.3cm t\in[0,1],
\end{align} 
where $P_{S_a}$ and $P_{\bar S}$ are the cumulative distribution functions of $S_a$ and $\bar S$ \cite{peyre2019computational}. This geodesic, by its uniqueness, is therefore the shortest path. The parameter $t$ controls the level to which $p_{S_a}$ is moved toward the barycenter $p_{\bar S}$, with $t=1$ corresponding to total matching.

An implementation of this method is described in Algorithm~\ref{Alg:wass2_pp_tradeoff}, where $\mathcal{D}_a$ denotes the subset of $\mathcal{D}$ corresponding to the group of $N_a$ individuals with sensitive attributes $a$. 
\subsection{General Fair Empirical Risk Minimization}
\label{sec:GFERM}
In Sect. \ref{sec:CDOT}, we discussed a post-processing method that was derived within a unified optimal transport approach to fairness. In this section, we present another unified framework for fairness, introduced in \cite{donini2018empirical,oneto2019general}, based on the \emph{empirical risk minimization} strategy. This framework incorporates several notions of fairness, can deal with continuous outputs and sensitive attributes through discretization, and is accompanied by risk and fairness bounds, which imply consistency properties both in terms of fairness measure and risk of the selected model. 

Let $\mathcal{D} = \{ (s^n,x^n,y^n) \}_{n=1}^N$ be a training dataset formed by $N$ samples drawn independently from an unknown probability distribution $\mu$ over $\mathcal{S} \times \mathcal{X} \times  \mathcal{Y}$, where $y^n$ is outcome that we wish to predict, $s^n$ the sensitive attribute, and $x^n$ a vector of features to be used to form a prediction $\hat y^n$ of $y^n$. Notice that we indicate the sensitive attribute with $s^n$ rather than $a^n$ as in the previous sections---we will use this new notation in all remaining sections.

To deal with the case in which $y^n$ and $s^n$ are continuous, we define the discretization sets $\mathcal{Y}_K = \{ t_1, {\cdots}, t_{K+1} \} \subset \mathbb{R}$ and $\mathcal{S}_Q = \{ \sigma_1, {\cdots}, \sigma_{Q+1} \} \subset \mathbb{R}$, where $t_1 < t_2 < \cdots < t_{K+1}$, $\sigma_1 < \sigma_2 < \cdots < \sigma_{Q+1}$, and $K$ and $Q$ are positive integers. The sets $\mathcal{Y}_K$ and $\mathcal{S}_{Q}$ define values of the outcome and sensitive attribute that are regarded as indistinguishable---their definition is driven by the particular application under consideration.
We indicate with $\mathcal{D}_{k,q}$ the subset of $N_{k,q}$ individuals with $y^n \in [t_k, t_{k+1})$ and  $s^n \in [\sigma_q, \sigma_{q+1})$ for $1\leq k \leq K$ and $1\leq q \leq Q$.

Unlike the previous sections, we indicate the model output with $f(z^n)$, where $f$ is a deterministic function (we refer to it as model) chosen from a set $\mathcal{F}$ such that $f: \mathcal{Z} \rightarrow \mathbb{R}$ ,
where $\mathcal{Z} = \mathcal{S} \times \mathcal{X}$ or $\mathcal{Z} = \mathcal{X}$, i.e. $\mathcal{Z}$ may contain or not the sensitive attribute. 

The \emph{risk of the model}, $L(f)$, is defined as $L(f) = \mathbb{E} \left[ \ell(f(Z),Y) \right]$, where $\ell:\mathbb{R} \times \mathcal{Y} \rightarrow \mathbb{R}$ is a \emph{loss function}. When necessary, we indicate with a subscript the particular loss function used and the associated risk, e.g.~$L_p(f) = \mathbb{E} \left[ \ell_p(f(Z),Y) \right]$.

We aim at minimizing the risk subject to the $\epsilon$-Loss General Fairness constraint introduced below, which generalizes previously known notions of fairness, and encompasses both classification and regression and categorical and numerical sensitive attributes.
\begin{definition}
\label{4.2:def:fairnessGF}
A model $f$ is \emph{$\epsilon$-General Fair} ($\epsilon$-GF) if it satisfies 
\begin{align}
\frac{1}{K Q^2}
\sum_{k = 1}^{K} \sum_{p,q = 1}^{Q} \left| P^{k,p}(f) - P^{k,q}(f)\right| \leq {\epsilon}, ~~~\epsilon \in [0,1],
\end{align}
where 
\begin{align}
P^{k,q}(f) = \mathbb{P} \big( f(Z) \in [t_k, t_{k+1}) ~| ~ Y \in [t_k, t_{k+1}), S \in [\sigma_q, \sigma_{q+1}) \big).
\end{align}
\end{definition}
This definition considers a model as fair if its predictions are approximately (with $\epsilon$ corresponding to the amount of acceptable approximation) equally distributed independently of the value of the sensitive attribute. It can be further generalized as follows.
\begin{definition}
\label{4.2:def:fairnessLGF}
A model $f$ is \emph{$\epsilon$-Loss General Fair} ($\epsilon$-LGF) if it satisfies 
\begin{align}
\frac{1}{K Q^2}
\sum_{k = 1}^{K} \sum_{p,q = 1}^Q \left| L^{k,p}_k(f) - L^{k,q}_k(f) \right| \leq \epsilon, ~~~\epsilon \in [0,1],
\end{align}
where 
\begin{align}
L^{k,q}_k(f) = \mathbb{E} \big[ \ell_k(f(Z),Y) ~ | ~Y \in [t_k, t_{k+1}), S \in [\sigma_q, \sigma_{q+1}) \big],
\end{align}
where $\ell_k$ is a loss function.
\end{definition}
This definition considers a model as fair if its errors, relative to the loss function,
are approximately equally distributed independently of the value of the sensitive attribute. 
\begin{remark}
For $ \ell_k(f(Z),Y) = \1_{f(Z)\notin[t_k, t_{k+1})}$, Definition~\ref{4.2:def:fairnessLGF} becomes Definition~\ref{4.2:def:fairnessGF}.
Moreover, it is possible to link Definition~\ref{4.2:def:fairnessLGF} to other fairness definitions in the literature.
Let us consider the setting $\mathcal{Y} = \{-1, +1 \}$, $\mathcal{S} = \{0, 1 \}$, $\mathcal{Y}_K = \{-1.5,$ $0,$ $ +1.5 \}$, $\mathcal{S}_Q = \{-0.5,$ $ 0.5,$ $ 1.5 \}$, $\epsilon = 0$. In this setting, if $\ell_k$ is the 0-1-loss, i.e. $\ell_k(f(Z),Y) = \1_{f(Z) Y \leq 0}$, then Definition~\ref{4.2:def:fairnessLGF} reduces to EFPRs/EFNRs (see Sect. \ref{sec:CBNs:VT}), whilst if $\ell_k$ is the linear loss, i.e. $\ell_k(f(Z),Y) = (1 - f(Z)Y)/2$, then we recover other notions of fairness introduced in~\cite{dwork2018decoupled}.
In the setting $\mathcal{Y} \subseteq \mathbb{R}$, $\mathcal{S} = \{0, 1 \}$, $\mathcal{Y}_K = \{-\infty, \infty \}$, $\mathcal{S}_Q =$ $ \{-0.5,$ $0.5,$ $1.5 \}$, $\epsilon = 0$, Definition~\ref{4.2:def:fairnessLGF} reduces to the notion of Mean
Distance introduced in~\cite{calders2013controlling} and also exploited in~\cite{komiyama2017two}.
Finally, in the same setting, if $\mathcal{S} \subseteq \mathbb{R}$ in~\cite{komiyama2017two} it is proposed to use the correlation coefficient which is equivalent to setting $\mathcal{S}_Q = \mathcal{S}$ in Definition~\ref{4.2:def:fairnessLGF}.
\end{remark}
Minimizing the risk subject to the $\epsilon$-LGF constraint corresponds to the following minimization problem
\begin{align}
\min_{f \in \mathcal{F}} \left\{ L(f) : 
\sum_{k = 1}^{K} \sum_{p,q = 1}^{Q} \left| L^{k,p}_k(f) - L^{k,q}_k(f) \right| \leq \epsilon
\right\},~~~\epsilon \in [0,1].
\label{4.2:eq:alg:deterministic}
\end{align}
Since $\mu$ is usually unknown and therefore the risks cannot be computed, we approximate this problem by minimizing the empirical 
counterpart 
\begin{align}
\min_{f \in \mathcal{F}} \left\{ \hat{L}(f) :
\sum_{k = 1}^{K} \sum_{p,q = 1}^{Q} \left| \hat{L}^{k,p}_k(f) - \hat{L}^{k,q}_k(f) \right| \leq {\hat \epsilon}
\right\},~~~\hat{\epsilon} \in [0,1],
\label{4.2:eq:alg:empirical}
\end{align}
where $\hat{L}(f) =  \mathbb{\hat E} \left[ \ell(f(Z),Y) \right]= \frac{1}{N} \sum_{(z^n,y^n) \in \mathcal{D}} \ell(f(z^n),y^n)$ and $\hat{L}^{k,q}_k(f) = \frac{1}{N_{k,q}} \sum_{(z^n,y^n) \in \mathcal{D}_{k,q}} \ell_k(f(z^n),y^n)$. We refer to~Problem~\eqref{4.2:eq:alg:empirical} as \emph{General Fair Empirical Risk Minimization} (G-FERM) since it generalizes the Fair Empirical Risk Minimization approach introduced in~\cite{donini2018empirical}.
\paragraph{{\bf Statistical Analysis}}
In this section we show that, if the parameter $\hat{\epsilon}$ is chosen appropriately, 
a solution $\hat{f}$ of Problem~\eqref{4.2:eq:alg:empirical} is in a certain sense a consistent estimator 
for a solution $f^*$ of Problems~\eqref{4.2:eq:alg:deterministic}. For this purpose we require that, for any data distribution, it holds with probability at least $1-\delta$ with respect to the draw of a dataset that
\begin{align}
\sup_{f \in \mathcal{F}} \big|L(f) - \hat{L}(f)\big| \leq B(\delta,N,\mathcal{F}),
\label{4.2:eq:bartlett}
\end{align}
where $B(\delta,N,\mathcal{F})$ goes to zero as $N$ grows to infinity, i.e. the class $\mathcal{F}$ is learnable with respect to the loss~\cite{shalev2014understanding}.
Moreover $B(\delta,N,\mathcal{F})$ is usually an exponential bound, which means that $B(\delta,N,\mathcal{F})$ grows logarithmically with respect to the inverse of $\delta$.

\begin{remark}
\label{4.2:rem:2}
If $\mathcal{F}$ is a compact subset of linear separators in a reproducing kernel Hilbert space, and the loss is Lipschitz in its first argument, then $B(\delta,N,\mathcal{F})$ can be obtained via Rademacher bounds~\cite{bartlett2002rademacher}.
In this case $B(\delta,N,\mathcal{F})$ goes to zero at least as ${\sqrt{1/N}}$ as $N$ grows and decreases with $\delta$ as ${\sqrt{\ln\left(1/\delta\right)}}$.
\end{remark}
We are ready to state the first result of this section.
\begin{theorem}
\label{thm:GFERMmainresult1}
Let $\mathcal{F}$ be a learnable set of functions with respect to the loss function $\ell: \mathbb{R} \times {\cal Y} \rightarrow \mathbb{R}$, and let $f^*$ and $\hat{f}$ be a solution of Problems~(\ref{4.2:eq:alg:deterministic}) and ~(\ref{4.2:eq:alg:empirical}) respectively, with 
\begin{align}
\hat{\epsilon} = \epsilon + \sum_{k = 1}^{K} \sum_{q,q' = 1}^{Q} 
\sum_{p \in \{q,q'\}} 
B(\delta,N_{k,p},\mathcal{F}).
\end{align}
With probability at least $1- \delta$ it holds simultaneously that
\begin{align}
\hskip-0.2cm \sum_{k = 1}^{K} \sum_{p,q = 1}^{Q} \left| L^{k,p}_k(f) - L^{k,q}_k(f) \right| \leq \epsilon \!+\! 2 \sum_{k = 1}^{K} \sum_{q,q' = 1}^{Q} 
\sum_{p \in \{q,q'\}} 
\hskip-0.2cm { B\left(\frac{\delta}{(4 K Q^2 + 2)},N_{k,p},\mathcal{F}\right)}, \hskip-0.1cm
\label{eq:first}
\end{align}
and
\begin{align}
L(\hat{f}) - L(f^*) \leq 2
{ B\left(\frac{\delta}{(4 K Q^2 + 2)},N,\mathcal{F}\right)}.
\label{eq:second}
\end{align}
\end{theorem}
The proof is reported in~\cite{oneto2019general}. A consequence of the first statement in Theorem~\ref{thm:GFERMmainresult1} is that, as $N$ tends to infinity, $L(\hat{f})$ tends to a value which is not larger than $L(f^*)$, i.e. G-FERM is consistent with respect to the risk of the selected model.
The second statement in Theorem~\ref{thm:GFERMmainresult1} instead implies that, as $N$ tends to infinity, $\hat{f}$ tends to satisfy the fairness criterion. 
In other words, G-FERM is consistent with respect to the fairness of the selected model.
\begin{remark}
Since $K,Q \leq N$, the bound in Theorem~\ref{thm:GFERMmainresult1} behaves as $\sqrt{\ln\left(1/\delta\right)/N}$ in the same setting of Remark~\ref{4.2:rem:2} which is optimal~\cite{shalev2014understanding}.
\end{remark}
Thanks to Theorem~\ref{thm:GFERMmainresult1}, we can state that $f^{*}$ is close to $\hat{f}$ both in term of its risk and its fairness.
Nevertheless, the final goal is to find a $f^*_h$ which solves the following problem 
\begin{align}\label{4.2:eq:problemHard}
\min_{f \in \mathcal{F}} \left\{ {L}(f) : 
\sum_{k = 1}^{K} \sum_{p,q= 1}^{Q} \left| {P}^{k,p}(f) - {P}^{k,q}(f) \right| \leq \epsilon
\right\}.
\end{align}
The quantities in Problem~\eqref{4.2:eq:problemHard} cannot be computed since the underline data generating distribution is unknown.
Moreover, the objective function and the fairness constraint are non convex. Theorem~\ref{thm:GFERMmainresult1} allows us to solve the first issue since we can safely search for a solution $\hat{f}_h$ of the empirical counterpart of Problem~\eqref{4.2:eq:problemHard}, which is given by
\begin{align}\label{4.2:eq:problemHardempirical}
\min_{f \in \mathcal{F}}\left\{ \hat{L}(f) : 
\sum_{k = 1}^{K} \sum_{p,q = 1}^{Q} \left| \hat{P}^{k,p}(f) - \hat{P}^{k,q}(f) \right| \leq \hat{\epsilon}
\right\} ,
\end{align}
where 
\begin{align}
\hat{P}^{k,q}(f) =
\frac{1}{N_{k,q}} \sum_{(z^n,y^n) \in \mathcal{D}_{k,q}} \1_{ f(z^n) \in [t_k, t_{k+1})}.
\label{4.2:eq:phatlucapontil}
\end{align}
Unfortunately, Problem~\eqref{4.2:eq:problemHardempirical} is still a difficult non-convex non-smooth problem.
Therefore, we replace the possible non-convex loss function in the risk with its convex upper bound $\ell_c$ (e.g.~the square loss $\ell_{c}(f(Z),Y) = (f(Z)-Y)^2$ for regression, or the hinge loss $\ell_{c}(f(Z),Y)=\max(0,1-f(Z)Y)$ for binary classification~\cite{shalev2014understanding}), and the losses $\ell_{k}$ in the constraint with a relaxation (e.g.~the linear loss $\ell_l(f(Z),Y) = f(Z) - Y$) which allows to make the constraint convex. This way we look for a solution $\hat{f}_c$ of the convex G-FERM problem
\begin{align}\label{4.2:eq:problemSoft}
\min_{f \in \mathcal{F}} \left\{ \hat{L}_c(f) : 
\sum_{k = 1}^{K} \sum_{p, q = 1}^{Q} \left| \hat{L}_l^{k,p}(f) - \hat{L}_l^{k,q}(f) \right| \leq \hat{\epsilon} 
\right\} .
\end{align}
This approximation of the fairness constraint corresponds to matching the first order moments~\cite{donini2018empirical}. Methods that attempt to match all moments, such as the one discussed in Sect. \ref{sec:CDOT} or \cite{quadrianto2017recycling}, or the second order moments~\cite{woodworth2017learning} are preferable, but result in non-convex problems.

The questions that arise here are whether $\hat{f}_c$ is close to $\hat{f}_h$, how much, and under which assumptions.
The following proposition sheds some lights on these questions.
\begin{proposition}
\label{4.2:thm:mainresult2}
If $\ell_c$ is a convex upper bound of the loss exploited to compute the risk then $
\hat{L}_{h}(f) \leq \hat{L}_{c}(f)$.
Moreover, if for $f: \mathcal{X} \rightarrow \mathbb{R}$ and for $\ell_l$
\begin{align}
\sum_{k = 1}^{K} \sum_{p,q = 1}^{Q} \left| \hat{P}^{k,p}(f) - \hat{P}^{k,q}(f) \right| - \left| \hat{L}_l^{k,p}(f) - \hat{L}_l^{k,q}(f) \right| \leq \hat{\Delta},
\end{align}
with $\hat{\Delta}$ small, then also the fairness is well approximated.
\end{proposition}
The first statement of Proposition~\ref{4.2:thm:mainresult2} tells us that the quality of the risk approximation depends on the quality of the convex approximation.
The second statement of Proposition~\ref{4.2:thm:mainresult2}, instead, tells us that if $\hat{\Delta}$ is small then the linear loss based fairness is close to $\epsilon$-LGF.
This condition is quite natural, empirically verifiable, and it has been exploited in previous work~\cite{donini2018empirical,maurer2004note}.
Moreover, in~\cite{oneto2019general} the authors present experiments showing that $\hat{\Delta}$ is small.

The bound in Proposition~\ref{4.2:thm:mainresult2} may be tighten by using different non-linear approximations of $\epsilon$-LGF.
However, the linear approximation proposed here gives a convex problem and, as showed in~\cite{oneto2019general}, works well in practice.

In summary, Theorem~\ref{thm:GFERMmainresult1} and Proposition~\ref{4.2:thm:mainresult2} give the conditions under which a solution $\hat{f}_c$ of Problem~\eqref{4.2:eq:alg:empirical}, which is convex, is close, both in terms of risk and fairness measure, to a solution $f^*_h$ of Problem~\eqref{4.2:eq:problemHard}.
\subsubsection{General Fair Empirical Risk Minimization with Kernel Methods}
\label{sec:GFERM-K}
In this section, we introduce a specific method for the case in which the underlying space of models is a reproducing kernel Hilbert space (RKHS)~\cite{shawe2004kernel,smola2001learning}.

Let $\mathbb{H}$ be the Hilbert space of square summable sequences, $\kappa: \mathcal{Z} \times \mathcal{Z} \rightarrow \mathbb{R}$ a positive definite kernel, and $\phi : \mathcal{Z} \rightarrow \mathbb{H}$ an induced feature mapping such that $\kappa(Z,\bar Z) = \langle \phi(Z),\phi(\bar Z)\rangle$, for all $Z,\bar Z \in \mathcal{Z}$. 
Functions $f$ in the RKHS can be parametrized as 
\begin{equation}
f(Z) = \langle w , \phi(Z)\rangle,~~~Z \in \mathcal{Z},
\label{4.2:eq:222}
\end{equation}
for some vector of parameters $w \in \mathbb{H}$.
Whilst a bias term can be added to $f$, we do not include it here for simplicity of exposition.

We propose to solve Problem~\eqref{4.2:eq:problemSoft} for the case in which $\mathcal{F}$ is a ball in the RKHS, using a convex loss function $\ell_{c}(f(Z),Y)$ to measure the empirical error and a linear loss function $\ell_l$ as fairness constraint.
We introduce the mean of the feature vectors associated with the training points restricted by the discretization of the sensitive attribute and real outputs, namely
\begin{align}
u_{k,q} = \frac{1}{N_{k,q}} \sum_{ (z^n,y^n) \in \mathcal{D}_{k,q}} \phi(z^n).
\end{align}
Using Eq.~\eqref{4.2:eq:222}, the constraint in Problem~\eqref{4.2:eq:problemSoft} becomes
\begin{align}
\sum_{k = 1}^{K} \sum_{p,q = 1}^{Q} \left| \langle w,u_{k,p} -u_{k,q} \rangle \right| \leq \hat{\epsilon},
\end{align}
which can be written as $\|A^T w\|_1 \leq {\hat \epsilon}$, where $A$ is the matrix having as columns the vectors $u_{k,p} - u_{k,q}$.
With this notation, the fairness constraint can be interpreted as the composition of $\hat \epsilon$ 
ball of the $\ell_1$ norm with a linear transformation $A$. In practice, we solve the following Tikhonov regularization problem
\begin{align}
\min\limits_{w \in \mathbb{H}} 
\sum_{(z^n,y^n) \in \mathcal{D}}
\ell_c(\langle w , \phi(z^n)\rangle,y^n) + \lambda \|w\|^2, 
~~~~\text{s.t. }~ \|A^\top w\|_1 \leq {\hat \epsilon},
\label{4.2:prob:ker}
\end{align}
where $\lambda$ is a positive parameter.
Note that, if $\hat{\epsilon} = 0$, the constraint reduces to the linear constraint $A^\top w =0$. Problem~\eqref{4.2:prob:ker} can be kernelized by observing that, thanks to the Representer Theorem~\cite{shawe2004kernel}, $w =  \sum_{(z^n,y^n) \in \mathcal{D}} \phi(z^n)$.
The dual of Problem~\eqref{4.2:prob:ker} may be derived using Fenchel duality, see e.g.~\cite[Theorem 3.3.5]{borwein2010convex}.

Finally we notice that, in the case in which $\phi$ is the identity mapping (i.e.~$\kappa$ is the linear kernel on $\mathbb{R}^d$) and $\hat{\epsilon}=0$, the fairness constraint of Problem~\eqref{4.2:prob:ker} can be implicitly enforced by making a change of representation~\cite{donini2018empirical}.
\subsubsection{Fair Empirical Risk Minimization through Pre-Processing}
\label{sec:FERM}
In this section, we show how the in-processing G-FERM approach described above can be translated into a pre-processing approach. For simplicity of exposition, we focus on the case of binary outcome and sensitive attribute, i.e. $\mathcal{Y} = \{ -1, +1 \}$ and $\mathcal{S} = \{0,1\}$, and assume $\mathcal{Z} = \mathcal{X}$.
We denote with $\mathcal{D}_{+,s}$ the subset of $N_{+,s} $ individuals belonging to class $1$ and with sensitive attribute $s^n=s$.
As above, the purpose of a learning procedure is to find a model that minimizes the empirical risk $\hat{L}(f) = \hat{\mathbb{E}} [\ell(f(X),Y)]$.

Le us introduce a slightly less general notion of fairness with respect to Definition \ref{4.2:def:fairnessLGF} of Sect. \ref{sec:GFERM}.
\begin{definition}
\label{def:fairness}
A model $f$ is \emph{$\epsilon$-Fair} ($\epsilon$-F)  if it satisfies the condition $| {L}^{+,0}(f) - {L}^{+,1}(f)| \leq \epsilon$, where $\epsilon \in [0,1]$ and ${L}^{+,s}(f)$ is the risk of the positive labeled samples with sensitive attribute $s$. 
\end{definition}
We aim at minimizing the risk subject to the fairness constraint given by Definition~\ref{def:fairness} with $\ell_h(f(X),Y) = \1_{f(X) Y \leq 0}$ (corresponding to Equal False Positive Rates for $\epsilon = 0$). Specifically, we consider the problem
\begin{align}
 \min\Big\{L_h(f) : f \in \mathcal{F} ,~
\big| {L}^{+,0}_h(f) - {L}^{+,1}_h(f)\big| \leq \epsilon\Big\}.
\label{eq:alg:deterministic}
\end{align}
By replacing the deterministic quantity with their empirical counterparts, the hard loss in the risk with a convex loss function $\ell_c$, and the hard loss in the constraint with the linear loss $\ell_l$, we obtaining the convex problem
\begin{align}\label{eq:problemSoft}
\min\Big\{\hat{L}_c(f) : f \in \mathcal{F} ,~
\big| \hat{L}^{+,0}_l(f) - \hat{L}^{+,1}_l(f)\big| \leq \hat{\epsilon}\Big\}.
\end{align}
In the case in which the underlying space of models is a RKHS, $f$ can parametrized as 
\begin{equation}
f(X) = \langle w , \phi(X)\rangle,~~~X \in \mathcal{X}.
\label{eq:222}
\end{equation}
Let $u_s$ be the barycenter in the feature space of the positively labelled points with sensitive attribute $s$, i.e.
\begin{align}
u_s= \frac{1}{N_{+,s}} \sum_{ n \in \mathcal{N}_{+,s}}\phi(x^n),
\end{align}
where $\mathcal{N}_{+,s} = \{n: y^n = 1, s^n = s \}$.
Using Eq.~\eqref{eq:222}, Problem~\eqref{eq:problemSoft} with Tikhonov regularization and for the case in which $\mathcal{F}$ is a ball in the RKHS takes the form 
\begin{align}
\min\limits_{w \in \mathbb{H}} ~
\sum_{n=1}^N \ell_c(\langle w,\phi(x^n)\rangle ,y^n) + \lambda \|w\|^2,
~~~~\text{s.t. } ~ \big|\langle w,u\rangle\big| \leq \epsilon, 
\label{prob:ker} 
\end{align}
where $u = u_0 - u_1$, and $\lambda$ is a positive parameter which controls model complexity. 

Using the Representer Theorem and the fact that $u$ is a linear combination of the feature vectors (corresponding to the subset of positive labeled points), we obtain $w=\sum_{n=1}^N \alpha_n\phi(x^n)$, and therefore $f(X) = \sum_{n=1}^N \alpha_n \kappa(x^n,X)$.

Let $K$ be the Gram matrix. The vector of coefficients $\alpha$ can then be found by solving 
\begin{align}
\min_{\alpha \in \mathbb{R}^N} ~
& \sum_{i=1}^N \ell\bigg(\sum_{j=1}^N K_{ij}\alpha_j,y^i\bigg) + \lambda \sum_{i,j=1}^N\alpha_i \alpha_j K_{ij},~~~~
\text{s.t. } ~
\bigg| 
\sum_{i=1}^N \alpha_i
\bigg[ 
\frac{1}{N_{+,0}} \sum_{j \in \mathcal{N}_{+,0}} K_{ij} 
- 
\frac{1}{N_{+,1}} \sum_{j \in \mathcal{N}_{+,1}} K_{ij} 
\bigg] 
\bigg| \leq \epsilon .
\end{align}
When $\phi$ is the identity mapping (i.e.~$\kappa$ is the linear kernel on $\mathbb{R}^d$) and $\epsilon=0$, we can solve the constraint $\langle w,u\rangle = 0$ for $w_i$, where the 
index $i$ is such that $| u_i | = \|u\|_\infty$, obtaining $w_{i} = - \sum_{j=1, j \neq i}^d w_j \frac{u_j}{u_i}$. Consequently, the linear model rewrites as
$\sum_{j=1}^{d} w_j x_j = \sum_{j=1, j \neq i}^d w_j (x_j - x_i \frac{u_j}{u_i})$.
In other words, the fairness constraint is implicitly enforced by making the change of representation $x \mapsto \tilde{x} \in \mathbb{R}^{d-1}$, with
\begin{equation}
\tilde{x}_j = x_j - x_i \frac{u_j}{u_i}, \quad j \in \{ 1, \dots, i-1, i+1, \dots, d \},
\label{eq:gggg}
\end{equation}
that has one feature fewer than the original one. This approach can be extended to the non-linear case by defining a fair kernel matrix instead of fair data mapping~\cite{donini2018empirical,oneto2019general}.
\subsection{Learning Fair Representations from Multiple Tasks}
\label{sec:LFR}
Pre-Processing methods aim at transforming the training data to remove undesired biases, i.e., most often, to make it statistically independent of sensitive attributes. Most existing methods consider a fixed scenario with training dataset ${\mathcal D} = \{(s^n,x^n,y^n)\}_{n=1}^{N}$, and achieve a transformation of $x^n\in \mathbb{R}^d$ through a mapping $g: \mathbb{R}^d \rightarrow \mathbb{R}^r$ which is either problem independent or dependent. In the latter case, the model is consider as a composition $f(g(X))$, where $g$ synthesizes the information needed to solve a particular task by learning a function $f$. We refer to such a problem-dependent mapping as \emph{representation}, and to a representation $g$ such that $g(X)$ does not depend on sensitive attributes as \emph{fair representation}.

There exist several approaches to learning fair representations. The work in~\cite{beutel2017data,edwards2016censoring,louizos2016variational,madras2018learning,mcnamara2019costs,mcnamara2017provably,wang2018invariant} propose different neural networks architectures together with modified learning strategies to learn a representation $g(X)$ that preserves information about $X$, is useful for predicting $Y$, and is approximately independent of the sensitive attribute. In~\cite{johansson2016learning} the authors show how to formulate the problem of counterfactual inference as a domain adaptation problem---specifically a covariate shift problem~\cite{quionero2009dataset}---and derive two families of representation algorithms for counterfactual inference. In~\cite{zemel2013learning}, the authors learn a representation that is a probability distribution over clusters, where learning the cluster of a datapoint contains no-information about the sensitive attribute. 

In a large number of real world applications using the same model or part of it over different tasks might be desirable. For example, it is common to perform a fine tuning over pre-trained models~\cite{donahue2014decaf}, keeping the internal representation fixed. Unfortunately, fine tuning a model which is fair on a task on novel previously unseen tasks could lead to an unexpected unfairness behaviour (i.e.~discriminatory transfer~\cite{discriminative_transfer} or negative legacy~\cite{kamishima2012fairness}), due to missing generalization guarantees concerning its fairness properties.
To avoid this issue, it is necessary to consider the learning problem in a multitask/lifelong learning framework. Recent methods leverage task similarity to learn fair representations that provably generalizes well to unseen tasks. 

In this section, we present the method introduced in~\cite{oneto2019learning} to learn a shared fair representation from multiple tasks, where each task could be a binary classification or regression problem.

Let us indicate with $\tau_{t} = (s^n_{t},{x}^n_{t},y^n_{t})_{n=1}^{N}$ the training sequence for task $t$, sampled independently from a probability distribution $\mu_t$ on $\mathcal{S} \times \mathcal{X} \times \mathcal{Y}$. The goal is to learn a model $f_t: \mathcal{Z} \times \mathcal{S} \rightarrow \mathcal{Y}$ for several tasks $t\in \{1,\dots,T\}$. For simplicity, we assume linear $f_t$ and $\mathcal{Z}=\mathcal{X}$, i.e. $f_t(X) = \langle {w}_t, X \rangle$ where ${w}_t \in \mathbb{R}^d$ is a vector of parameters, and binary sensitive attributes $\mathcal{S} = \{0,1\}$---the method naturally extends to the non-linear and multiple sensitive attribute cases.

Following a general \emph{multi-task learning} (MTL) formulation, we aim at minimizing the multitask empirical error plus a regularization term which leverages similarities between the tasks.
A natural choice for the regularizer is given by the trace norm, namely the sum of the singular values of the matrix $W = [{w}_1 \cdots {w}_T] \in \mathbb{R}^{d\times T}$. This results in the following matrix factorization problem
\begin{align}
\min_{A,B} \quad & \frac{1}{TN}\sum_{t=1}^T \sum_{n=1}^{N} \left( y^n_{t}-\langle {b}_t, A\trans{x}^n_{t}\rangle \right)^2 +\frac{\lambda}{2} \left(\|A\|_F^2 + \|B\|_F^2\right),
\label{eq:2}
\end{align}
where $A=[a_1\dots a_r] \in \mathbb{R}^{d \times r}$, $B =[b_1\dots b_T]\in \mathbb{R}^{r \times T}$ with $W=AB$, and where $\| \cdot \|_F$ is the Frobenius norm (see e.g.~\cite{srebro2004learning} and references therein).
Here $r \in \mathbb{N}$ is the number of factors, i.e. the upper bound on the rank of $W$.
If $r \geq \min(d,T)$, Problem~\eqref{eq:2} is equivalent to trace norm regularization~\cite{argyriou2008convex} (see e.g.~\cite{ciliberto2017reexamining} and references therein\footnote{If $r <\min(d,T)$, Problem~\eqref{eq:2} is equivalent to trace norm regularization plus a rank constraint.}).
Problem \eqref{eq:2} can be solved by gradient descent or alternate minimization as we discuss next.
Once the problem is solved, the estimated parameters of the function $w_t$ for the tasks' linear models are simply computed as $w_t=Ab_t$.
Notice that the problem is stated with the square loss function for simplicity, but the observations extend to the general case of proper convex loss functions.

The approach can be interpreted as learning a two-layer network with linear activation functions.
Indeed, the matrix $A\trans$ applied to $X$ induces the linear representation $A\trans X = ( {a}_1\trans X, \cdots, {a}_r\trans X)\trans$.
We would like each component of the representation vector to be independent of the sensitive attribute on each task.
This means that, for every measurable subset $C \subset \mathbb{R}^{r}$ and for every $t \in \{1,\dots,T\}$, we would like
\begin{align}
\label{eq:comp}
\mathbb{P}( A\trans X_t \in C~ |~ S = 0) = \mathbb{P}(A\trans X_t \in C ~|~ S = 1).
\end{align}
To turn the non-convex constraint into a convex one, we require only the means to be the same, and compute those from empirical data.
For each training sequence $\tau \in ({\cal X} \times {\cal Y})^T$, we define the empirical conditional means
\begin{align}
c({\tau}) = \frac{1}{|\mathcal{N}_0(\tau)|} \sum_{n\in \mathcal{N}_{0}(\tau)} {x}^n - 
\frac{1}{|\mathcal{N}_1(\tau)|} \sum_{n\in \mathcal{N}_{1}(\tau)} {x}^n,
\end{align}
where $\mathcal{N}_{s}(\tau) = \{n: s^n = s \}$, and relax the constraint of Eq.~\eqref{eq:comp} to
\begin{align}
\label{eq:epsT}
A\trans {c}({\tau}_t)
= 0.
\end{align}
This give the following optimization problem
\begin{align}
\label{prob:mt}
\min_{A,B} \quad & \frac{1}{TN}\sum_{t=1}^T \sum_{n=1}^{N} \left( y^n_{t}-\langle b_t, A\trans{x}^n_{t}\rangle \right)^2 +\frac{\lambda}{2} \left(\|A\|_F^2 + \|B\|_F^2 \right) \\
& A\trans {c}({\tau}_t)
= 0, \quad
t \in \{1,\dots,T\}.
\nonumber
\end{align}
We tackle Problem~\eqref{prob:mt} with alternate minimization. Let ${y}_t = [y^1_{t}, \dots, y^N_{t}]\trans$ be the vector formed by the outputs of task $t$, and let $X_t = [({x}^1_{t})\trans,\dots, ({x}^n_{t})\trans]\trans$ be the data matrix for task $t$. When we regard $A$ as fixed and solve w.r.t.
$B$, Problem~\eqref{prob:mt} can be reformulated as
\begin{align}
\min_{B} \quad & 
\left\| 
\begin{bmatrix}
{y}_1 \\
\vdots\\
{y}_T \\
\end{bmatrix}
- 
\begin{bmatrix}
X_1 A & 0 & \cdots & 0 \\
\vdots & \vdots & \vdots & \vdots \\
0 & \cdots & 0 & X_T A \\
\end{bmatrix}
\begin{bmatrix}
{b}_1 \\
\vdots\\
{b}_T \\
\end{bmatrix}
\right\|^2 
+ \lambda 
\left\| 
\begin{bmatrix}
{b}_1 \\
\vdots \\
{b}_T \\
\end{bmatrix}
\right\|^2 ,
\end{align}
which can be easily solved.
In particular, notice that the problem decouples across tasks, and each task specific problem amounts to running ridge regression on the data transformed by the representation matrix $A\trans$.
When instead $B$ is fixed and we solve w.r.t.
$A$, Problem~\eqref{prob:mt} can be reformulated as
\begin{align}
\min_{A} \quad & 
\left\| 
\begin{bmatrix}
{y}_1 \\
\vdots\\
{y}_T \\
\end{bmatrix}
- 
\begin{bmatrix}
b_{1,1} X_1 & \cdots & b_{1,r} X_1 \\
& \vdots \\
b_{t,1} X_T & \cdots & b_{t,r} X_T \\
\end{bmatrix} 
\begin{bmatrix}
{a}_1 \\
\vdots \\
{a}_r \\
\end{bmatrix} 
\right\|^2 
+ \lambda 
\left\| 
\begin{bmatrix}
{a}_1 \\
\vdots \\
{a}_r \\
\end{bmatrix}
\right\|^2,  
~~~~\text{s.t. } ~
\begin{bmatrix}
{a}_1^T \\
\vdots \\
{a}_r^T \\
\end{bmatrix}
\circ
\begin{bmatrix}
{c}_1, \dots, {c}_T
\end{bmatrix} = {0},
\end{align} 
where we used the shorthand notation $c_t = c({\tau}_t)$, and where $\circ$ is the Kronecker product for partitioned tensors (or Tracy-Singh product).
Consequently, by alternating minimization we can solve the original problem.
Notice also that we may relax the equality constraint as $\frac{1}{T}\sum_{t=1}^T \|A\trans c({\tau}_t)\|^2 \leq \epsilon$, where $\epsilon$ is some tolerance parameter.
In fact, this may be required when the vectors $c({\tau}_t)$ span the entire input space. In this case, we may also add a soft constraint in the regularizer.

Notice that, if independence is satisfied at the representation level, i.e. if Eq.~\eqref{eq:comp} holds, then every model built from such a representation will satisfies independence at the output level.
Likewise, if the representation satisfies the convex relaxation (Eq.~\eqref{eq:epsT}), then it also holds that $\langle w_t,c({\tau}_t)\rangle = \langle b_t, A\trans c({\tau}_t)\rangle =0$, i.e. the task weight vectors satisfy the first order moment approximation.
More importantly, as we show below, if the tasks are randomly observed, then independence (or its relaxation) will also be satisfied on future tasks with high probability.
In this sense, the method can be interpreted as learning a fair transferable representation.
\subsubsection*{Learning Bound}
In this section we study the learning ability of the method proposed above.
We consider the setting of learning-to-learn~\cite{baxter2000model}, in which the training tasks (and their corresponding datasets) used to find a fair data representation are regarded as random variables from a meta-distribution.
The learned representation matrix $A$ is then transferred to a novel task, by applying ridge regression on the task dataset, in which $X$ is transformed as $A\trans X$.
In~\cite{maurer2009transfer} a learning bound is presented, linking the average risk of the method over tasks from the meta-distribution (the so-called transfer risk) to the multi-task empirical error on the training tasks.
This result quantifies the good performance of the representation learning method when the number of tasks grow and the data distribution on the raw input data is intrinsically high dimensional (hence learning is difficult without representation learning).
We extend this analysis to the setting of algorithmic fairness, in which the performance of the algorithm is evaluated both relative to risk and the fairness constraint.
We show that both quantities can be bounded by their empirical counterparts evaluated on the training tasks.

Let ${\cal E}_\mu(w) = \mathbb{E}_{(X,Y) \sim \mu} [(Y-\langle w,X \rangle )^2]$ and ${\cal E}_{\tau}(w) = \frac{1}{N} \sum_{n=1}^N (y^n-\langle w,x^n\rangle)^2$. For every matrix $A \in \mathbb{R}^{d \times r}$ and for every data sample $\tau = (x^n,y^n)_{n=1}^N$, let $b_A(\tau)$ be the minimizer of ridge regression with modified data representation, i.e. $b_A(\tau)= \arg\min_{b \in \mathbb{R}^r} \frac{1}{N} \sum_{n=1}^N (y^n-\langle b,A\trans x^n\rangle)^2 + \lambda \|b\|^2$. 
\begin{theorem}
\label{prop:1}
Let $A$ be the representation learned by solving Problem~\eqref{eq:2} and renormalized so that $\|A\|_F = 1$.
Let tasks $\mu_1,\dots,\mu_T$ be independently sampled from a meta-distribution $\rho$, and let $z_t$ be sampled from $\mu_t$ for $t\in\{1,\cdots,T\}$.
Assume that the input marginal distribution of random tasks from $\rho$ is supported on the unit sphere and that the outputs are in the interval $[-1,1]$, almost surely.
Let $r = \min(d,T)$.
Then, for any $\delta\in(0,1]$ it holds with probability at least $1-\delta$ in the drawing of the datasets ${\tau}_1,\dots,{\tau}_T$, that
\begin{align}
 \mathbb{E}_{\mu\sim\rho}\mathbb{E}_{\tau\sim\mu}& ~ {\cal R}_\mu\big(w_A(\tau)\big) - \frac{1}{T} \sum_{t=1}^T {\cal R}_{\tau_t}(w_A(\tau_t)) \nonumber \\
& \leq \frac{4}{\lambda}\sqrt{\frac{ \|{\hat C}\|_\infty}{N}} + \frac{24}{\lambda N}\sqrt{\frac{ \ln \frac{8 NT}{\delta}}{T}} + \frac{14}{\lambda} \sqrt{\frac{\ln(NT)\|{\hat C\|_\infty}}{T}} + \sqrt{\frac{2\ln \frac{4}{\delta} }{T}},
\end{align}
and
\begin{equation}
\mathbb{E}_{\mu\sim\rho}\mathbb{E}_{{\mathcal \tau\sim\mu}} \| A c(\tau)\|^2 - \frac{1}{T} \sum_{t=1}^T\| A c(\tau_t) \|^2 \leq 96 \frac{ \ln \frac{8r^2}{\delta}}{T}+ 6 \sqrt{\frac{{\|\hat \Sigma}\|_\infty \ln \frac{8 r^2}{\delta}}{T}}.
\end{equation}
\end{theorem}
The proof is reported in~\cite{oneto2019learning}.

Notice that the first bound in Theorem~\ref{prop:1} improves Theorem~2 in~\cite{maurer2009transfer}.
The improvement is due to the introduction of the empirical total covariance in the second term in the right-hand side of the inequality.
The result in~\cite{maurer2009transfer} instead contains the term $\sqrt{1/T}$, which can be considerably larger when the raw input is distributed on a high dimensional manifold.
The bounds in Theorem~\ref{prop:1} can be extended to hold with variable sample size per task.
In order to simplify the presentation, we assumed that all datasets are composed of the same number of datapoints $N$.
The general setting can be addressed by letting the sample size be a random variable and introducing the slightly different definition of the transfer risk in which we also take the expectation w.r.t.
the sample size.
The hyperparameter $\lambda$ is regarded as fixed in the analysis.
In practice it is chosen by cross-validation.
The bound on fairness measure contains two terms in the right-hand side, in the spirit of Bernstein's inequality.
The slow term $O(1/\sqrt{T})$ contains the spectral norm of the covariance of difference of means across the sensitive groups.
Notice that $\|\Sigma\|_\infty \leq 1$, but it can be much smaller when the means are close to each other, i.e. when the original representation is already approximately fair.
\subsection{If the Explicit Use of Sensitive Attributes is Forbidden}
\label{sec:FairModels:Sens}
Due to legal requirements, developing methods that work without explicit use of sensitive attributes is a central problem in ML fairness. 

The criterion \emph{Fairness through Unawareness} was introduced to capture the legal requirement of not explicitly using sensitive attributes to form decisions. This criterion states that a model output $\hat Y$ is fair as long as it does not make explicit use of the sensitive attribute $S$. However, from a modeling perspective, not explicitly using $S$ can result in a less accurate model, without necessarily improving the fairness of the solution \cite{dwork2018decoupled,pedreshi2008discrimination,zafar2017fairness}. This could be the case if some variables used to form $\hat Y$ depend on $S$. CBNs give us the instruments to understand that there might be even a more subtle issue with this fairness criterion, 
as explained in the following example introduced in \cite{kusner2017counterfactual} and discussed in \cite{chiappa2019causal}.

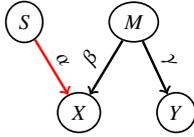
\begin{wrapfigure}[6]{l}{0.22\textwidth}
\vskip-0.4cm
\centering
\scalebox{0.8}{
\begin{tikzpicture}[dgraph]
\node[ocont] (A) at (-0.9,1.5) {$S$};
\node[ocont] (M) at (0.9,1.5) {$M$};
\node[ocont] (X) at (0,0) {$X$};
\node[ocont] (Y) at (1.6,0) {$Y$};
\draw[line width=1.15pt,red](A)--node[sloped,above,black]{$\alpha$}++(X);
\draw[line width=1.15pt](M)--node[sloped,above]{$\beta$}++(X);
\draw[line width=1.15pt](M)--node[sloped,above]{$\gamma$}++(Y);
\end{tikzpicture}}
\vspace{-0.4cm}
\end{wrapfigure}

\noindent Consider the CBN on the left representing the data-generation mechanism underlying a music degree scenario, where $S$ corresponds to gender, $M$ to music aptitude (unobserved, i.e.~$M\notin \mathcal{D}$), $X$ to the score obtained from an ability test taken at the beginning of the degree, and $Y$ to the score obtained from an ability test taken at the end of the degree.

\noindent Individuals with higher music aptitude $M$ are more likely to obtain higher initial and final scores ($M\rightarrow X$, $M\rightarrow Y$).
Due to discrimination occurring at the initial testing, women are assigned a lower initial score than men for the same aptitude level ($S \rightarrow X$).
The only path from $S$ to $Y$, $S\rightarrow X \leftarrow M \rightarrow Y$, is closed as $X$ is a \emph{collider} on this path.
Therefore the unfair influence of $S$ on $X$ does not reach $Y$ ($Y\independent S$).
Nevertheless, as $Y\cancel{\independent}S |X$, a prediction $\hat Y$ based only on the initial score $X$ would contain the unfair influence of $S$ on $X$.
For example, assume the following linear model
\begin{align}
Y=\gamma M, ~~ X =\alpha S + \beta M, ~~~~\text{with } ~\mathbb{E}_{p(S)}[S^2]=1, ~\mathbb{E}_{p(M)}[M^2]=1.
\end{align}
A linear predictor of the form $\hat Y = \theta_X X$ minimizing $\mathbb{E}_{p(S)p(M)}[(Y-\hat Y)^2]$ would have parameters $\theta_X=\gamma\beta/(\alpha^2+\beta^2)$, giving $\hat Y = \gamma\beta(\alpha S + \beta M)/(\alpha^2+\beta^2)$, i.e.~$\hat Y\cancel{\independent} S$.
Therefore, this predictor would be using the sensitive attribute to form a decision, although implicitly rather than explicitly.
Instead, a predictor explicitly using the sensitive attribute, $\hat Y = \theta_X X + \theta_S S$, would have parameters
\begin{align}
\left(\begin{array}{c}
\theta_X \\
\theta_S \\
\end{array} \right)
&=\left(
\begin{array}{cc}
\alpha^2+\beta^2 & \alpha \\
\alpha & 1 \\
\end{array} \right)^{-1}
\left(\begin{array}{c}
\gamma\beta \\
0\\
\end{array} \right)
=\left(\begin{array}{c}
\gamma/\beta \\
-\alpha\gamma/\beta \\
\end{array} \right),
\end{align}
i.e.~$\hat Y = \gamma M$.
Therefore, this predictor would be fair. In general (e.g.~in a non-linear setting) it is not guaranteed that using $S$ would ensure $\hat Y \independent S$.
Nevertheless, this example shows how explicit use of $S$ in a model can ensure fairness rather than leading to unfairness. In summary, not being able to explicit use sensitive attributes to build a model might be problematic for several reasons. 

Less strict legal requirements prohibit the explicit use of $S$ when deploying the model, but permit it during its training (see~\cite{dwork2018decoupled} and references therein). This case can be dealt with using different approaches. A simple approach (used e.g. in \cite{jiang2019wasserstein}) would be to use explicitly the sensitive attribute only in the constraint term used to enforce fairness. This, however, might be suboptimal for similar reasons to not using the sensitive attribute at all. 

In this section, we describe a different approach introduced in \cite{oneto2019taking}. In this approach, a function $g: \mathcal{X} \rightarrow \mathcal{S}$ that forms a prediction, $\hat{S} = g(X)$, of the sensitive attribute $S$ from $X$ is learned. Then $\hat{S}$, instead of $S$, is used to learn group specific models via a multi-task learning approach (MTL). 
As shown in \cite{oneto2019taking}, if the prediction $\hat S$ is accurate, this approach allows to exploit MTL to learn group specific models. Instead, if $\hat S$ is inaccurate, this approach acts as a randomization procedure which improves the fairness measure of the overall model.

We focus on binary outcome and on categorical sensitive attribute, i.e. $\mathcal{Y} = \{ -1, +1 \}$ and  $\mathcal{S} = \{1, \cdots , k \}$. Using the operator $\diamond \in \{-,+\}$, we denote with $\mathcal{D}_{\diamond,s}$, the subset of $N_{\diamond,s} $ individuals belonging to class $\diamond$ and with sensitive attribute $s$. 
We consider the case in which the underlying space of models is a RKHS, leading to the functional form
\begin{equation}
f(S,X) =  w \cdot \phi(S,X),~~~(S,X) \in \mathcal{S} \times \mathcal{X},
\end{equation}
where $``\cdot"$ is the inner product between two vectors in a Hilbert space\footnote{For all intents and purposes, one may also assume throughout that $\mathbb{H}=\mathbb{R}^d$, the standard $d$-dimensional vector space, for some positive integer $d$.}. We can then learn the parameter vector $w$ by $\| w \|^2$-regularized empirical risk minimization. 

The average accuracy with respect to each group of a model $L(f)$, together with its empirical counterparts $\hat{L}(f)$, are defined respectively as
\begin{align}
L(f) = \frac{1}{k} \sum_{s \in \mathcal{S}} L_s(f), \quad 
L_s(f) = \mathbb{E} \left[\ell(f(S=s,X),Y) \right],\quad s \in \mathcal{S},
\end{align}
and
\begin{align}
& \hat{L}(f) = \frac{1}{k} \sum_{s \in \mathcal{S}} \hat{L}_s(f), \quad 
\hat{L}_s(f) = \frac{1}{N_s} \sum_{(x^n,s^n,y^n) \in \mathcal{D}_s} \ell(f(s^n,x^n),y^n), \quad s \in \mathcal{S},
\end{align}
where $\ell:\mathbb{R} \times \mathcal{Y} \rightarrow \mathbb{R}$ is the error loss function.
In the following, we first briefly discuss the MTL approach, and then explain how it can be enhanced with fairness constraints.\\

%
We use a multi-task learning approach based on regularization around a common mean~\cite{evgeniou2004regularized}. We choose $\phi(S,X) = (0_{s-1},\varphi(X),0_{k-s},\varphi(X))$, so that $f(S,X) = w_0 \cdot \varphi(X) + v_S \cdot \varphi(X)$ for $w_0, v_S \in \mathbb{H}$. MTL jointly learns a shared model $w_0$ and task specific models $w_s = w_0 + v_s$ $\forall s \in \mathcal{S}$ by encouraging them to be close to each other. This is achieved with the following Tikhonov regularization problem
\begin{align}
\label{4.4:eq:MTL}
& \min_{
\tiny
\begin{matrix}
w_0,
\cdots, 
w_S 
\end{matrix}
} 
\theta \hat{L}(w_0) + (1-\theta) \frac{1}{k} 
\sum_{s =1}^k \hat{L}_s(w_s)
+ \rho \left[\lambda \|w_0\|^2 + (1-\lambda) 
\frac{1} {k}
\sum_{s=1}^k \| w_s - w_0 \|^2\right],
\end{align}
where the parameter $\lambda \in [0,1]$ encourages closeness between the shared and specific models, and the parameter $\theta \in [0,1]$ captures the relative importance of the loss of the shared model and the group-specific models. The MTL problem is convex provided that the loss function used to measure the empirical errors ${\hat L}$ and ${\hat L}_s$ in~\eqref{4.4:eq:MTL} are convex.\\

As fairness criterion, we consider EFPRs/EFNRs (Equalized Odds)
%
\begin{align}\label{4.4:eq:1}
\mathbb{P}\{f(S,X) > 0 | S = 1, Y= {\diamond}  1\} = \cdots = \mathbb{P}\{f(S,X) > 0 | S = k, Y= {\diamond}  1\}.
\end{align}
In many recent papers~\cite{adebayo2016iterative,agarwal2018reductions,alabi2018optimizing,bechavod2018Penalizing,berk2017convex,beutel2017data,calmon2017optimized,donini2018empirical,dwork2018decoupled,feldman2015certifying,hardt2016equality,kamiran2009classifying,kamiran2010classification,kamiran2012data,kamishima2011fairness,kearns2018preventing,menon2018cost,perez-suay2017fair,pleiss2017fairness,woodworth2017learning,zafar2017fairness,zafar2017fairnessARXIV,zafar2017parity,zemel2013learning} it has been shown how to enforce EFPRs/EFNRs during model training.
Here we build upon the approach proposed in~\cite{donini2018empirical} since it is convex, theoretically grounded, and performs favorably against state-of-the-art alternatives.
To this end, we first observe that
\begin{align}\label{4.4:eq:3}
\!\!\!\!\mathbb{P}\{f(S,X) > 0\ | \ S = s, Y= \diamond 1 \} = 1 - \mathbb{E} [ \ell_h(f(S=s,X),Y=\diamond 1 )]  = 1 - L_s(f),
\end{align}
where $\ell_h(f(S,X),Y) = \1_{f(S,X)Y\leq 0}$.
Then, by substituting Eq.~\eqref{4.4:eq:3} in Eq.~\eqref{4.4:eq:1}, replacing the deterministic quantities with their empirical counterpart, and by approximating the hard loss function $\ell_h$ with the linear one $\ell_l = (1-f(S,X)Y)/2$ we obtain the convex constraint 
\begin{align}\label{4.4:eq:4}
\frac{1}{N_{\diamond,1}} \sum_{(s^n,x^n,y^n) \in \mathcal{D}_{\diamond,1}} f(s^n,x^n) = \cdots = \frac{1}{N_{\diamond,k}} \sum_{(s^n,x^n,y^n) \in \mathcal{D}_{\diamond,k}} f(s^n,x^n).
\end{align}
Enforcing this constraint can achieved by adding to the MTL the $(k-1)$ constraints 
\begin{align}
w_1 \cdot u^\diamond_1 = w_2 \cdot u^\diamond_2
\ \wedge \ \cdots \ \wedge \
w_1 \cdot u^\diamond_1 = w_k \cdot u^\diamond_k,
\end{align}
where $u^\diamond_s = \frac{1}{N_{\diamond,s}} \sum_{ (s^n,x^n) \in \mathcal{D}_{\diamond,s}}
\varphi(s^n,x^n)$. Thanks to the Representer Theorem, as shown in~\cite{donini2018empirical}, it is straightforward to derive a kernelized version of the MTL convex problem which can be solved with any solver.
\section{Conclusions}
\label{sec:Conclusions}
In this manuscript, we have discussed an emerging area of machine learning that studies the development of techniques for ensuring that models do not treat individuals unfairly due to biases in the data and model inaccuracies. 
Rather than an exhaustive descriptions of existing fairness criteria and approaches to impose fairness in a model, we focused on highlighting a few important points and research areas that we believe should get more attention in the literature, and described some of the work that we have done in these areas.

In particular, we have demonstrated that CBNs provide us with a precious tool to reason about and deal with fairness. A common criticism to CBNs is that the true underlying data-generation mechanism, and therefore the graph structure, is seldom known.
However, our discussion has demonstrated that this framework can nevertheless be helpful for reasoning at a high level about fairness and to avoid pitfalls. 
As unfairness in a dataset often displays complex patterns, we need a way to characterize and account for this complexity: CBNs represents the best currently available tool for achieving that. Furthermore, imposing criteria such as path-specific fairness, whilst difficult in practise, is needed to address the arguably most common unfairness scenarios present in the real world. 

We have described an optimal transport approach to fairness that enables us to account for the full shapes of distributions corresponding to different sensitive attributes, overcoming the limitations of most current methods that approximate fairness desiderata by considering lower order moments, typically the first moments, of those distributions. More research is needed in this direction.

In modern contexts, models are often not learned from scratch for solving new tasks, since datasets are too complex or small in cardinality. To ensure that fairness properties generalize to multiple tasks, it is necessary to consider the learning problem in a multitask/lifelong learning framework. We described a method to learn fair representations that can generalize to unseen task.

Finally, we have discussed legal restrictions with the use of sensitive attributes, and introduced an in-processing approach that does not require the use of sensitive attributes during the deployment of the model. 

A limitation of this manuscript is that it does not discuss the important aspect that decisions have consequences on the future of individuals, and therefore fairness should be considered also in a temporal, rather than static, setting. 
\section*{Acknowledgments}
This work was partially supported by Amazon AWS Machine Learning Research Award.
\bibliographystyle{plain}
\bibliography{main.bib}

\begin{thebibliography}{100}

\bibitem{adebayo2016iterative}
J.~Adebayo and L.~Kagal.
\newblock Iterative orthogonal feature projection for diagnosing bias in
  black-box models.
\newblock In {\em Fairness, Accountability, and Transparency in Machine
  Learning}, 2016.

\bibitem{adler2018auditing}
P.~Adler, C~Falk, S.~A. Friedler, T.~Nix, G.~Rybeck, C.~Scheidegger, B.~Smith,
  and S.~Venkatasubramanian.
\newblock Auditing black-box models for indirect influence.
\newblock {\em Knowledge and Information Systems}, 54(1):95--122, 2018.

\bibitem{agarwal2018reductions}
A.~Agarwal, A.~Beygelzimer, M.~Dudik, J.~Langford, and H.~Wallach.
\newblock A reductions approach to fair classification.
\newblock In {\em Proceedings of the 35th International Conference on Machine
  Learning}, pages 60--69, 2018.

\bibitem{ainowinstitute2018litigating}
{AI Now Institute}.
\newblock Litigating algorithms: Challenging government use of algorithmic
  decision systems, 2016.

\bibitem{alabi2018unleashing}
D.~Alabi, N.~Immorlica, and A.~T. Kalai.
\newblock Unleashing linear optimizers for group-fair learning and
  optimization.
\newblock In {\em 31st Annual Conference on Learning Theory}, pages 2043--2066,
  2018.

\bibitem{alabi2018optimizing}
D.~Alabi, N.~Immorlica, and A.~T. Kalai.
\newblock When optimizing nonlinear objectives is no harder than linear
  objectives.
\newblock {\em CoRR}, abs/1804.04503, 2018.

\bibitem{ali2019loss}
J.~Ali, M.~B. Zafar, A.~Singla, and K.~P. Gummadi.
\newblock Loss-aversively fair classification.
\newblock In {\em Proceedings of the 2019 AAAI/ACM Conference on AI, Ethics,
  and Society}, pages 211--218, 2019.

\bibitem{amrieh2015students}
E.~A. Amrieh, T.~Hamtini, and I.~Aljarah.
\newblock {Students' Academic Performance Data Set}.
\newblock Available at \url{https://www.kaggle.com/aljarah/xAPI-Edu-Data},
  2015.

\bibitem{OnetoC003}
D.~Anguita, A.~Ghio, L.~Oneto, and S.~Ridella.
\newblock Selecting the hypothesis space for improving the generalization
  ability of support vector machines.
\newblock In {\em IEEE International Joint Conference on Neural Networks},
  2011.

\bibitem{angwin2016machine}
J.~Angwin, J.~Larson, S.~Mattu, and L.~Kirchner.
\newblock {Machine Bias: There's software used across the country to predict
  future criminals. And it's biased against blacks.}, 2016.

\bibitem{argyriou2008convex}
A.~Argyriou, T.~Evgeniou, and M.~Pontil.
\newblock Convex multi-task feature learning.
\newblock {\em Machine Learning}, 73(3):243--272, 2008.

\bibitem{bartlett2002rademacher}
P.~L. Bartlett and S.~Mendelson.
\newblock Rademacher and {G}aussian complexities: Risk bounds and structural
  results.
\newblock {\em Journal of Machine Learning Research}, 3:463--482, 2002.

\bibitem{baxter2000model}
J.~Baxter.
\newblock A model of inductive bias learning.
\newblock {\em Journal of Artificial Intelligence Research}, 12:149--198, 2000.

\bibitem{bechavod2018Penalizing}
Y.~Bechavod and K.~Ligett.
\newblock Penalizing unfairness in binary classification.
\newblock {\em CoRR}, abs/1707.00044, 2018.

\bibitem{berk2017convex}
R.~Berk, H.~Heidari, S.~Jabbari, M.~Joseph, M.~Kearns, J.~Morgenstern, S.~Neel,
  and A.~Roth.
\newblock A convex framework for fair regression.
\newblock In {\em Fairness, Accountability, and Transparency in Machine
  Learning}, 2017.

\bibitem{beutel2017data}
A.~Beutel, J.~Chen, Z.~Zhao, and E.~H. Chi.
\newblock Data decisions and theoretical implications when adversarially
  learning fair representations.
\newblock {\em CoRR}, abs/1707.00075, 2017.

\bibitem{bogen2018hiring}
M.~Bogen and A.~Rieke.
\newblock Help wanted: An examination of hiring algorithms, equity, and bias.
\newblock Technical report, Upturn, 2018.

\bibitem{borwein2010convex}
J.~Borwein and A.~S. Lewis.
\newblock {\em Convex Analysis and Nonlinear Optimization: Theory and
  Examples}.
\newblock Springer, 2010.

\bibitem{bureau2019national}
{Bureau of Labor Statistics}.
\newblock {National Longitudinal Surveys Of Youth Data Set}.
\newblock Available at \url{https://www.bls.gov/nls/}, 2019.

\bibitem{byanjankar2015predicting}
A.~Byanjankar, M.~Heikkil{\"a}, and J.~Mezei.
\newblock Predicting credit risk in peer-to-peer lending: A neural network
  approach.
\newblock In {\em IEEE Symposium Series on Computational Intelligence}, 2015.

\bibitem{calders2009building}
T.~Calders, F.~Kamiran, and M.~Pechenizkiy.
\newblock Building classifiers with independency constraints.
\newblock In {\em Data Mining Workshops, 2009. ICDMW'09. IEEE International
  Conference on}, pages 13--18, 2009.

\bibitem{calders2013controlling}
T.~Calders, A.~Karim, F.~Kamiran, W.~Ali, and X.~Zhang.
\newblock Controlling attribute effect in linear regression.
\newblock In {\em IEEE International Conference on Data Mining}, 2013.

\bibitem{calders2010three}
T.~Calders and S.~Verwer.
\newblock Three naive bayes approaches for discrimination-free classification.
\newblock {\em Data Mining and Knowledge Discovery}, 21(2):277--292, 2010.

\bibitem{calmon2017optimized}
F.~Calmon, D.~Wei, B.~Vinzamuri, K.~N. Ramamurthy, and K.~R. Varshney.
\newblock Optimized pre-processing for discrimination prevention.
\newblock In {\em Proceedings of the 31st Conference on Neural Information
  Processing Systems}, pages 3995--4004, 2017.

\bibitem{chiappa2019path}
S.~Chiappa.
\newblock Path-specific counterfactual fairness.
\newblock In {\em Thirty-Third AAAI Conference on Artificial Intelligence},
  pages 7801--7808, 2019.

\bibitem{chiappa2019causal}
S.~Chiappa and W.~S. Isaac.
\newblock A causal {B}ayesian networks viewpoint on fairness.
\newblock In {\em E. Kosta, J. Pierson, D. Slamanig, S. Fischer-H{\"u}bner, S.
  Krenn (eds) Privacy and Identity Management. Fairness, Accountability, and
  Transparency in the Age of Big Data. Privacy and Identity 2018. IFIP Advances
  in Information and Communication Technology}, volume 547. Springer, Cham,
  2019.

\bibitem{chiappa2020general}
S.~Chiappa, R.~Jiang, T.~Stepleton, A.~Pacchiano, H.~Jiang, and J.~Aslanides.
\newblock A general approach to fairness with optimal transport.
\newblock In {\em Thirty-Fourth AAAI Conference on Artificial Intelligence},
  2020.

\bibitem{chierichetti2017fair}
F.~Chierichetti, R.~Kumar, S.~Lattanzi, and S.~Vassilvitskii.
\newblock Fair clustering through fairlets.
\newblock In {\em Proceedings of the 31st Conference on Neural Information
  Processing Systems}, pages 5036--5044, 2017.

\bibitem{chouldechova17fair}
A.~Chouldechova.
\newblock Fair prediction with disparate impact: A study of bias in recidivism
  prediction instruments.
\newblock {\em Big Data}, 5(2):153--163, 2017.

\bibitem{chouldechova2018child}
A.~Chouldechova, E.~Putnam-Hornstein, D.~Benavides-Prado, O.~Fialko, and
  R.~Vaithianathan.
\newblock A case study of algorithm-assisted decision making in child
  maltreatment hotline screening decisions.
\newblock In {\em Proceedings of the 1st Conference on Fairness, Accountability
  and Transparency}, pages 134--148, 2018.

\bibitem{chzhen2019leveraging}
E.~Chzhen, H.~Hebiri, C.~Denis, L.~Oneto, and M.~Pontil.
\newblock Leveraging labeled and unlabeled data for consistent fair binary
  classification.
\newblock In {\em Proceedings of the 33rd Conference on Neural Information
  Processing Systems}, pages 12739--12750, 2019.

\bibitem{ciliberto2017reexamining}
C.~Ciliberto, D.~Stamos, and M.~Pontil.
\newblock Reexamining low rank matrix factorization for trace norm
  regularization.
\newblock {\em CoRR}, abs/1706.08934, 2017.

\bibitem{OnetoJ021}
L.~Coraddu, A.~Oneto, F.~Baldi, and D.~Anguita.
\newblock Vessels fuel consumption forecast and trim optimisation: a data
  analytics perspective.
\newblock {\em Ocean Engineering}, 130:351--370, 2017.

\bibitem{corbett-davies2017algorithmic}
S.~Corbett-Davies, E.~Pierson, A.~Feller, S.~Goel, and A.~Huq.
\newblock Algorithmic decision making and the cost of fairness.
\newblock In {\em ACM SIGKDD International Conference on Knowledge Discovery
  and Data Mining}, pages 797--806, 2017.

\bibitem{cortez2009wine}
P.~Cortez.
\newblock {Wine Quality Data Set}.
\newblock Available at
  \url{https://archive.ics.uci.edu/ml/datasets/Wine+Quality}, 2009.

\bibitem{cortez2014student}
P.~Cortez.
\newblock {Student Performance Data Set}.
\newblock Available at
  \url{https://archive.ics.uci.edu/ml/datasets/Student+Performance}, 2014.

\bibitem{cotter2018training}
A.~Cotter, M.~Gupta, H.~Jiang, N.~Srebro, K.~Sridharan, S.~Wang, B.~Woodworth,
  and S.~You.
\newblock Training well-generalizing classifiers for fairness metrics and other
  data-dependent constraints.
\newblock {\em CoRR}, abs/1807.00028, 2018.

\bibitem{cotter2019two}
A.~Cotter, H.~Jiang, and K.~Sridharan.
\newblock Two-player games for efficient non-convex constrained optimization.
\newblock In {\em Algorithmic Learning Theory}, 2019.

\bibitem{dawid2007fundamentals}
P.~Dawid.
\newblock {\em Fundamentals of Statistical Causality}.
\newblock Technical report, 2007.

\bibitem{defauw2018clinically}
J.~De~Fauw, J.~R. Ledsam, B.~Romera-Paredes, S.~Nikolov, N.~Tomasev,
  S.~Blackwell, H.~Askham, X.~Glorot, B.~O'Donoghue, D.~Visentin, G.~Van
  Den~Driessche, B.~Lakshminarayanan, C.~Meyer, F.~Mackinder, S.~Bouton,
  K.~Ayoub, R.~Chopra, D.~King, A.~Karthikesalingam, C.~O. Hughes, R.~Raine,
  J.~Hughes, D.~A. Sim, C.~Egan, A.~Tufail, H.~Montgomery, D.~Hassabis,
  G.~Rees, T.~Back, P.~T. Khaw, M.~Suleyman, J.~Cornebise, P.~A. Keane, and
  O.~Ronneberger.
\newblock Clinically applicable deep learning for diagnosis and referral in
  retinal disease.
\newblock {\em Nature Medicine}, 24(9):1342--1350, 2018.

\bibitem{dieterich2016compas}
W.~Dieterich, C.~Mendoza, and T.~Brennan.
\newblock {COMPAS} risk scales: Demonstrating accuracy equity and predictive
  parity, 2016.

\bibitem{doherty2012information}
Neil~A Doherty, Anastasia~V Kartasheva, and Richard~D Phillips.
\newblock Information effect of entry into credit ratings market: The case of
  insurers' ratings.
\newblock {\em Journal of Financial Economics}, 106(2):308--330, 2012.

\bibitem{donahue2014decaf}
J.~Donahue, Y.~Jia, O.~Vinyals, J.~Hoffman, N.~Zhang, E.~Tzeng, and T.~Darrell.
\newblock Decaf: A deep convolutional activation feature for generic visual
  recognition.
\newblock In {\em Proceedings of the 31st International Conference on Machine
  Learning}, pages 647--655, 2014.

\bibitem{donini2018empirical}
M.~Donini, L.~Oneto, S.~Ben-David, J.~S. Shawe-Taylor, and M.~Pontil.
\newblock Empirical risk minimization under fairness constraints.
\newblock In {\em Proceedings of the 32nd Conference on Neural Information
  Processing Systems}, pages 2791--2801, 2018.

\bibitem{dwork12fairness}
C.~Dwork, M.~Hardt, T.~Pitassi, O.~Reingold, and R.~Zemel.
\newblock Fairness through awareness.
\newblock In {\em Innovations in Theoretical Computer Science Conference},
  2012.

\bibitem{dwork2018decoupled}
C.~Dwork, N.~Immorlica, A.~T. Kalai, and M.~D.~M. Leiserson.
\newblock Decoupled classifiers for group-fair and efficient machine learning.
\newblock In {\em Proceedings of the 1st Conference on Fairness, Accountability
  and Transparency}, pages 119--133, 2018.

\bibitem{edwards2016censoring}
H.~Edwards and A.~Storkey.
\newblock Censoring representations with an adversary.
\newblock In {\em 4th International Conference on Learning Representations},
  2015.

\bibitem{eubanks2018automating}
V.~Eubanks.
\newblock {\em Automating Inequality: How High-Tech Tools Profile, Police, and
  Punish the Poor}.
\newblock St. Martin's Press, 2018.

\bibitem{evgeniou2004regularized}
T.~Evgeniou and M.~Pontil.
\newblock Regularized multi-task learning.
\newblock In {\em Proceedings of the 10th ACM SIGKDD International Conference
  on Knowledge Discovery and Data Mining}, pages 109--117, 2004.

\bibitem{fehrman2016drug}
E.~Fehrman, V.~Egan, and E.~M. Mirkes.
\newblock {Drug Consumption Data Set}.
\newblock Available at
  \url{https://archive.ics.uci.edu/ml/datasets/Drug+consumption+\%28quantified\%29},
  2016.

\bibitem{feldman2015computational}
M.~Feldman.
\newblock Computational fairness: Preventing machine-learned discrimination,
  2015.

\bibitem{feldman2015certifying}
M.~Feldman, S.~A. Friedler, J.~Moeller, C.~Scheidegger, and
  S.~Venkatasubramanian.
\newblock Certifying and removing disparate impact.
\newblock In {\em Proceedings of the 21th ACM SIGKDD International Conference
  on Knowledge Discovery and Data Mining}, pages 259--268, 2015.

\bibitem{fish2015fair}
B.~Fish, J.~Kun, and A.~Lelkes.
\newblock Fair boosting: a case study.
\newblock In {\em Fairness, Accountability, and Transparency in Machine
  Learning}, 2015.

\bibitem{fish2016confidence}
B.~Fish, J.~Kun, and A.~D. Lelkes.
\newblock A confidence-based approach for balancing fairness and accuracy.
\newblock In {\em SIAM International Conference on Data Mining}, pages
  144--152, 2016.

\bibitem{fitzsimons2018equality}
J.~Fitzsimons, A.~A. Ali, M.~Osborne, and S.~Roberts.
\newblock Equality constrained decision trees: For the algorithmic enforcement
  of group fairness.
\newblock {\em CoRR}, abs/1810.05041, 2018.

\bibitem{fukuchi2015prediction}
K.~Fukuchi, T.~Kamishima, and J.~Sakuma.
\newblock Prediction with model-based neutrality.
\newblock {\em IEICE TRANSACTIONS on Information and Systems},
  98(8):1503--1516, 2015.

\bibitem{gajane2017formalizing}
P.~Gajane and M.~Pechenizkiy.
\newblock On formalizing fairness in prediction with machine learning.
\newblock {\em CoRR}, abs/1710.03184, 2017.

\bibitem{gillen2018online}
S.~Gillen, C.~Jung, M.~Kearns, and A.~Roth.
\newblock Online learning with an unknown fairness metric.
\newblock In {\em Proceedings of the 32nd Neural Information Processing
  Systems}, pages 2600--2609, 2018.

\bibitem{goh2016satisfying}
G.~Goh, A.~Cotter, M.~Gupta, and M.~P. Friedlander.
\newblock Satisfying real-world goals with dataset constraints.
\newblock In {\em Proceedings of the 30th Conference on Neural Information
  Processing Systems}, pages 2415--2423, 2016.

\bibitem{goldstein1987school}
H.~Goldstein.
\newblock School effectiveness data set.
\newblock Available at
  \url{http://www.bristol.ac.uk/cmm/learning/support/datasets/}, 1987.

\bibitem{gordaliza2019obtaining}
P.~Gordaliza, E.~Del~Barrio, G.~Fabrice, and L.~Jean-Michel.
\newblock Obtaining fairness using optimal transport theory.
\newblock In {\em Proceedings of the 36th International Conference on
  International Conference on Machine Learning}, pages 2357--2365, 2019.

\bibitem{grgic2017fairness}
N.~Grgi{\'c}-Hla{\v{c}}a, M.~B. Zafar, K.~P. Gummadi, and A.~Weller.
\newblock On fairness, diversity and randomness in algorithmic decision making.
\newblock {\em CoRR}, abs/1706.10208, 2017.

\bibitem{guvenir1998arrhythmia}
H.~A. Guvenir, B.~Acar, and H.~Muderrisoglu.
\newblock {Arrhythmia Data Set}.
\newblock Available at
  \url{https://archive.ics.uci.edu/ml/datasets/Arrhythmia}, 1998.

\bibitem{hajian2012methodology}
S.~Hajian and J.~Domingo-Ferrer.
\newblock A methodology for direct and indirect discrimination prevention in
  data mining.
\newblock {\em IEEE Transactions on Knowledge and Data Engineering},
  25(7):1445--1459, 2012.

\bibitem{hajian2014generalization}
S.~Hajian, J.~Domingo-Ferrer, and O.~Farr{\`a}s.
\newblock Generalization-based privacy preservation and discrimination
  prevention in data publishing and mining.
\newblock {\em Data Mining and Knowledge Discovery}, 28(5-6):1158--1188, 2014.

\bibitem{hajian2011rule}
S.~Hajian, J.~Domingo-Ferrer, and A.~Martinez-Balleste.
\newblock Rule protection for indirect discrimination prevention in data
  mining.
\newblock In {\em International Conference on Modeling Decisions for Artificial
  Intelligence}, 2011.

\bibitem{hajian2015discrimination}
S.~Hajian, J.~Domingo-Ferrer, A.~Monreale, D.~Pedreschi, and F.~Giannotti.
\newblock Discrimination-and privacy-aware patterns.
\newblock {\em Data Mining and Knowledge Discovery}, 29(6):1733--1782, 2015.

\bibitem{hajian2012injecting}
S.~Hajian, A.~Monreale, D.~Pedreschi, J.~Domingo-Ferrer, and F.~Giannotti.
\newblock Injecting discrimination and privacy awareness into pattern
  discovery.
\newblock In {\em IEEE International Conference on Data Mining Workshops},
  2012.

\bibitem{hardt2016equality}
M.~Hardt, E.~Price, and N.~Srebro.
\newblock Equality of opportunity in supervised learning.
\newblock In {\em Proceedings of the 30th Conference on Neural Information
  Processing Systems}, pages 3315--3323, 2016.

\bibitem{harper2016movielens}
F.~M. Harper and J.~A. Konstan.
\newblock {Movielens Data Set}.
\newblock Available at \url{https://grouplens.org/datasets/movielens/}, 2016.

\bibitem{hashimoto2018fairness}
T.~B. Hashimoto, M.~Srivastava, H.~Namkoong, and P.~Liang.
\newblock Fairness without demographics in repeated loss minimization.
\newblock In {\em Proceedings of the 35th International Conference on on
  Machine Learning}, pages 1929--1938, 2018.

\bibitem{he2014practical}
X.~He, J.~Pan, O.~Jin, T.~Xu, B.~Liu, T.~Xu, Y.~Shi, A.~Atallah, R.~Herbrich,
  S.~Bowers, and J.~Q. Candela.
\newblock Practical lessons from predicting clicks on ads at facebook.
\newblock In {\em International Workshop on Data Mining for Online
  Advertising}, 2014.

\bibitem{hebert2017calibration}
U.~H{\'e}bert-Johnson, M.~P. Kim, O.~Reingold, and G.~N. Rothblum.
\newblock Calibration for the (computationally-identifiable) masses.
\newblock {\em CoRR}, abs/1711.08513, 2017.

\bibitem{heidari2018fairness}
H.~Heidari, C.~Ferrari, K.~Gummadi, and A.~Krause.
\newblock Fairness behind a veil of ignorance: A welfare analysis for automated
  decision making.
\newblock In {\em Proceedings of the 32nd Conference on Neural Information
  Processing Systems}, pages 1273--1283, 2018.

\bibitem{heidari2018moral}
H.~Heidari, M.~Loi, K.~P. Gummadi, and A.~Krause.
\newblock A moral framework for understanding of fair ml through economic
  models of equality of opportunity.
\newblock {\em CoRR}, abs/1809.03400, 2018.

\bibitem{heritage2011heritage}
{Heritage Provider Network}.
\newblock {Heritage Health Data Set}.
\newblock Available at \url{https://www.kaggle.com/c/hhp/data}, 2011.

\bibitem{hoffman2018discretion}
M.~Hoffman, L.~B. Kahn, and D.~Li.
\newblock Discretion in hiring.
\newblock {\em The Quarterly Journal of Economics}, 133(2):765--800, 2018.

\bibitem{hofmann1994statlog}
H.~Hofmann.
\newblock {Statlog (German Credit) Data Set}.
\newblock Available at
  \url{https://archive.ics.uci.edu/ml/datasets/statlog+(german+credit+data)},
  1994.

\bibitem{hu2019fair}
L.~Hu and Y.~Chen.
\newblock Fair classification and social welfare.
\newblock {\em CoRR}, abs/1905.00147, 2019.

\bibitem{hussain2018student}
S.~Hussain, N.~A. Dahan, F.~M. Ba-Alwib, and N.~Ribata.
\newblock {Student Academics Performance Data Set}.
\newblock Available at
  \url{https://archive.ics.uci.edu/ml/datasets/Student+Academics+Performance},
  2018.

\bibitem{isaac2017hope}
W.~S. Isaac.
\newblock Hope, hype, and fear: The promise and potential pitfalls of
  artificial intelligence in criminal justice.
\newblock {\em Ohio State Journal of Criminal Law}, 15(2):543--558, 2017.

\bibitem{jabbari2017fairness}
S.~Jabbari, M.~Joseph, M.~Kearns, J.~Morgenstern, and A.~Roth.
\newblock Fairness in reinforcement learning.
\newblock In {\em Proceedings of the 34th International Conference on Machine
  Learning}, pages 1617--1626, 2017.

\bibitem{janosi1988heart}
A.~Janosi, W.~Steinbrunn, M.~Pfisterer, and R.~Detrano.
\newblock {Heart Disease Data Set}.
\newblock Available at
  \url{https://archive.ics.uci.edu/ml/datasets/Heart+Disease}, 1988.

\bibitem{jiang2019wasserstein}
R.~Jiang, A.~Pacchiano, T.~Stepleton, H.~Jiang, and S.~Chiappa.
\newblock Wasserstein fair classification.
\newblock In {\em Thirty-Fifth Uncertainty in Artificial Intelligence
  Conference}, 2019.

\bibitem{johansson2016learning}
F.~Johansson, U.~Shalit, and D.~Sontag.
\newblock Learning representations for counterfactual inference.
\newblock In {\em Proceedings of The 33rd International Conference on Machine
  Learning}, pages 3020--3029, 2016.

\bibitem{johndrow2019algorithm}
J.~E. Johndrow and K.~Lum.
\newblock An algorithm for removing sensitive information: application to
  race-independent recidivism prediction.
\newblock {\em The Annals of Applied Statistics}, 13(1):189--220, 2019.

\bibitem{johnson2016impartial}
K.~D. Johnson, D.~P. Foster, and R.~A. Stine.
\newblock Impartial predictive modeling: Ensuring fairness in arbitrary models.
\newblock {\em CoRR}, abs/1608.00528, 2016.

\bibitem{joseph2016rawlsian}
M.~Joseph, M.~Kearns, J.~Morgenstern, S.~Neel, and A.~Roth.
\newblock Rawlsian fairness for machine learning.
\newblock In {\em Fairness, Accountability, and Transparency in Machine
  Learning}, 2016.

\bibitem{joseph2016fairness}
M.~Joseph, M.~Kearns, J.~H. Morgenstern, and A.~Roth.
\newblock Fairness in learning: Classic and contextual bandits.
\newblock In {\em Proceedings of the 30th Conference on Neural Information
  Processing Systems}, pages 325--333, 2016.

\bibitem{kamiran2009classifying}
F.~Kamiran and T.~Calders.
\newblock Classifying without discriminating.
\newblock In {\em International Conference on Computer, Control and
  Communication}, 2009.

\bibitem{kamiran2010classification}
F.~Kamiran and T.~Calders.
\newblock Classification with no discrimination by preferential sampling.
\newblock In {\em The Annual Machine Learning Conference of Belgium and The
  Netherlands}, 2010.

\bibitem{kamiran2012data}
F.~Kamiran and T.~Calders.
\newblock Data preprocessing techniques for classification without
  discrimination.
\newblock {\em Knowledge and Information Systems}, 33(1):1--33, 2012.

\bibitem{kamiran2012decision}
F.~Kamiran, A.~Karim, and X.~Zhang.
\newblock Decision theory for discrimination-aware classification.
\newblock In {\em IEEE International Conference on Data Mining}, 2012.

\bibitem{kamishima2012fairness}
T.~Kamishima, S.~Akaho, H.~Asoh, and J.~Sakuma.
\newblock Fairness-aware classifier with prejudice remover regularizer.
\newblock In {\em Joint European Conference on Machine Learning and Knowledge
  Discovery in Databases}, 2012.

\bibitem{kamishima2013independence}
T.~Kamishima, S.~Akaho, H.~Asoh, and J.~Sakuma.
\newblock The independence of fairness-aware classifiers.
\newblock In {\em IEEE International Conference on Data Mining Workshops},
  2013.

\bibitem{kamishima2011fairness}
T.~Kamishima, S.~Akaho, and J.~Sakuma.
\newblock Fairness-aware learning through regularization approach.
\newblock In {\em International Conference on Data Mining Workshops}, 2011.

\bibitem{kearns2018preventing}
M.~Kearns, S.~Neel, A.~Roth, and Z.~S. Wu.
\newblock Preventing fairness gerrymandering: Auditing and learning for
  subgroup fairness.
\newblock In {\em Proceedings of the 35th International Conference on Machine
  Learning}, pages 2564--2572, 2018.

\bibitem{kilbertus2017avoiding}
N.~Kilbertus, M.~R. Carulla, G.~Parascandolo, M.~Hardt, D.~Janzing, and
  B.~Sch{\"o}lkopf.
\newblock Avoiding discrimination through causal reasoning.
\newblock In {\em Proceedings of the 31th Conference on Neural Information
  Processing Systems}, pages 656--666, 2017.

\bibitem{kim2018fairness}
M.~Kim, O.~Reingold, and G.~Rothblum.
\newblock Fairness through computationally-bounded awareness.
\newblock In {\em Proceedings of the 32nd Conference on Neural Information
  Processing Systems}, pages 4842--4852, 2018.

\bibitem{kim2019multiaccuracy}
M.~P. Kim, A.~Ghorbani, and J.~Zou.
\newblock Multiaccuracy: Black-box post-processing for fairness in
  classification.
\newblock In {\em Proceedings of the 2019 AAAI/ACM Conference on AI, Ethics,
  and Society}, pages 247--254, 2019.

\bibitem{koepke2017danger}
J.~L. Koepke and D.~G. Robinson.
\newblock Danger ahead: Risk assessment and the future of bail reform.
\newblock {\em Washington Law Review}, 93:1725--1807, 2017.

\bibitem{kohavi1996census}
R.~Kohavi and B.~Becker.
\newblock {Census Income Data Set}.
\newblock Available at
  \url{https://archive.ics.uci.edu/ml/datasets/census+income}, 1996.

\bibitem{komiyama2017two}
J.~Komiyama and H.~Shimao.
\newblock Two-stage algorithm for fairness-aware machine learning.
\newblock {\em CoRR}, abs/1710.04924, 2017.

\bibitem{komiyama2018nonconvex}
J.~Komiyama, A.~Takeda, J.~Honda, and H.~Shimao.
\newblock Nonconvex optimization for regression with fairness constraints.
\newblock In {\em Proceedings of the 35th International Conference on Machine
  Learning}, pages 2737--2746, 2018.

\bibitem{kourou2015machine}
K.~Kourou, T.~P. Exarchos, K.~P. Exarchos, M.~V. Karamouzis, and D.~I.
  Fotiadis.
\newblock Machine learning applications in cancer prognosis and prediction.
\newblock {\em Computational and Structural Biotechnology Journal}, 13:8--17,
  2015.

\bibitem{kusner2017counterfactual}
M.~J. Kusner, J.~Loftus, C.~Russell, and R.~Silva.
\newblock Counterfactual fairness.
\newblock In {\em Proceedings of the 31st Conference on Neural Information
  Processing Systems}, pages 4069--4079, 2017.

\bibitem{discriminative_transfer}
C.~Lan and J.~Huan.
\newblock Discriminatory transfer.
\newblock {\em CoRR}, abs/1707.00780, 2017.

\bibitem{larson2016propublica}
J.~Larson, S.~Mattu, L.~Kirchner, and J.~Angwin.
\newblock {Propublica COMPAS Risk Assessment Data Set}.
\newblock Available at \url{https://github.com/propublica/compas-analysis},
  2016.

\bibitem{lim1997contraceptive}
T.~S. Lim.
\newblock {Contraceptive Method Choice Data Set}.
\newblock Available at
  \url{https://archive.ics.uci.edu/ml/datasets/Contraceptive+Method+Choice},
  1997.

\bibitem{lisini2007characterization}
S.~Lisini.
\newblock Characterization of absolutely continuous curves in {W}asserstein
  spaces.
\newblock {\em Calculus of Variations and Partial Differential Equations},
  28(1):85--120, 2007.

\bibitem{liu2015celeba}
Z.~Liu, P.~Luo, X.~Wang, and X.~Tang.
\newblock Celeb{A} {D}ata {S}et.
\newblock Available at \url{http://mmlab.ie.cuhk.edu.hk/projects/CelebA.html},
  2015.

\bibitem{louizos2016variational}
C.~Louizos, K.~Swersky, Y.~Li, M.~Welling, and R.~Zemel.
\newblock The variational fair autoencoder.
\newblock In {\em 4th International Conference on Learning Representations},
  2016.

\bibitem{lum2016predict}
K.~Lum and W.~S. Isaac.
\newblock To predict and serve?
\newblock {\em Significance}, 13(5):14--19, 2016.

\bibitem{lum2016statistical}
K.~Lum and J.~Johndrow.
\newblock A statistical framework for fair predictive algorithms.
\newblock {\em CoRR}, abs/1610.08077, 2016.

\bibitem{luo2015discrimination}
L.~Luo, W.~Liu, I.~Koprinska, and F.~Chen.
\newblock Discrimination-aware association rule mining for unbiased data
  analytics.
\newblock In {\em International Conference on Big Data Analytics and Knowledge
  Discovery}, pages 108--120. Springer, 2015.

\bibitem{luong2011k}
B.~T. Luong, S.~Ruggieri, and F.~Turini.
\newblock k-nn as an implementation of situation testing for discrimination
  discovery and prevention.
\newblock In {\em ACM SIGKDD International Conference on Knowledge Discovery
  and Data Mining}, 2011.

\bibitem{ma2015chicago}
D.~S. Ma, J.~Correll, and B.~Wittenbrink.
\newblock {Chicago Face Data Set}.
\newblock Available at \url{https://chicagofaces.org/default/}, 2015.

\bibitem{madras2018learning}
D.~Madras, E.~Creager, T.~Pitassi, and R.~Zemel.
\newblock Learning adversarially fair and transferable representations.
\newblock {\em CoRR}, abs/1802.06309, 2018.

\bibitem{madras2018predict}
D.~Madras, T.~Pitassi, and R.~Zemel.
\newblock Predict responsibly: improving fairness and accuracy by learning to
  defer.
\newblock In {\em Proceedings of the 32nd Conference on Neural Information
  Processing Systems}, pages 6147--6157, 2018.

\bibitem{malekipirbazari2015risk}
M.~Malekipirbazari and V.~Aksakalli.
\newblock Risk assessment in social lending via random forests.
\newblock {\em Expert Systems with Applications}, 42(10):4621--4631, 2015.

\bibitem{mancuhan2012discriminatory}
K.~Mancuhan and C.~Clifton.
\newblock Discriminatory decision policy aware classification.
\newblock In {\em IEEE International Conference on Data Mining Workshops},
  2012.

\bibitem{mancuhan2014combating}
K.~Mancuhan and C.~Clifton.
\newblock Combating discrimination using {B}ayesian networks.
\newblock {\em Artificial Intelligence and Law}, 22(2):211--238, 2014.

\bibitem{mary2019fairness}
J.~Mary, C.~Calauzenes, and N.~El~Karoui.
\newblock Fairness-aware learning for continuous attributes and treatments.
\newblock In {\em Proceedings of the 36th International Conference on Machine
  Learning}, pages 4382--4391, 2019.

\bibitem{maurer2004note}
A.~Maurer.
\newblock A note on the {PAC B}ayesian theorem.
\newblock {\em CoRR}, cs.LG/0411099, 2004.

\bibitem{maurer2009transfer}
A.~Maurer.
\newblock Transfer bounds for linear feature learning.
\newblock {\em Machine Learning}, 75(3):327--350, 2009.

\bibitem{mcnamara2017provably}
D.~McNamara, C.~S. Ong, and R.~C. Williamson.
\newblock Provably fair representations.
\newblock {\em CoRR}, abs/1710.04394, 2017.

\bibitem{mcnamara2019costs}
D.~McNamara, C.~Soon Ong, and B.~Williamson.
\newblock Costs and benefits of fair representation learning.
\newblock In {\em Proceedings of the 2019 AAAI/ACM Conference on AI, Ethics and
  Society}, pages 263--270, 2019.

\bibitem{menon2018cost}
A.~K. Menon and R.~C. Williamson.
\newblock The cost of fairness in binary classification.
\newblock In {\em Proceedings of the 1st Conference on Fairness, Accountability
  and Transparency}, pages 107--118, 2018.

\bibitem{merler2019diversity}
M.~Merler, N.~Ratha, R.~S. Feris, and J.~R. Smith.
\newblock {Diversity in Faces Data Set}.
\newblock Available at
  \url{https://research.ibm.com/artificial-intelligence/trusted-ai/diversity-in-faces/#highlights},
  2019.

\bibitem{mitchell2018fair}
S.~Mitchell, E.~Potash, and S.~Barocas.
\newblock Prediction-based decisions and fairness: A catalogue of choices,
  assumptions, and definitions.
\newblock {\em CoRR}, abs/1811.07867, 2018.

\bibitem{monge1781memoire}
G.~Monge.
\newblock {\em M{\'e}moire sur la th{\'e}orie des d{\'e}blais et des remblais}.
\newblock Histoire de l'Acad{\'e}mie Royale des Sciences de Paris, 1781.

\bibitem{moro2014bank}
S.~Moro, P.~Cortez, and P.~Rita.
\newblock {Bank Marketing Data Set}.
\newblock Available at
  \url{https://archive.ics.uci.edu/ml/datasets/bank+marketing}, 2014.

\bibitem{nabi2019learning}
R.~Nabi, D.~Malinsky, and I.~Shpitser.
\newblock Learning optimal fair policies.
\newblock In {\em Proceedings of the 36th International Conference on Machine
  Learning}, pages 4674--4682, 2019.

\bibitem{nabi2018fair}
R.~Nabi and I.~Shpitser.
\newblock Fair inference on outcomes.
\newblock In {\em Thirty-Second AAAI Conference on Artificial Intelligence},
  pages 1931--1940, 2018.

\bibitem{narasimhan2018learning}
H.~Narasimhan.
\newblock Learning with complex loss functions and constraints.
\newblock In {\em Proceedings of the Twenty-First International Conference on
  Artificial Intelligence and Statistics}, pages 1646--1654, 2018.

\bibitem{new2012stop}
{New York Police Department}.
\newblock {Stop, Question and Frisk Data Set}.
\newblock Available at
  \url{https://www1.nyc.gov/site/nypd/stats/reports-analysis/stopfrisk.page},
  2012.

\bibitem{noriega2019active}
Alejandro Noriega-Campero, Michiel~A Bakker, Bernardo Garcia-Bulle, and
  Alex'Sandy' Pentland.
\newblock Active fairness in algorithmic decision making.
\newblock In {\em Proceedings of the 2019 AAAI/ACM Conference on AI, Ethics,
  and Society}, pages 77--83, 2019.

\bibitem{olfat2017spectral}
M.~Olfat and A.~Aswani.
\newblock Spectral algorithms for computing fair support vector machines.
\newblock {\em CoRR}, abs/1710.05895, 2017.

\bibitem{oneto2019taking}
L.~Oneto, M.~Donini, A.~Elders, and M.~Pontil.
\newblock Taking advantage of multitask learning for fair classification.
\newblock In {\em AAAI/ACM Conference on AI, Ethics, and Society}, 2019.

\bibitem{oneto2019learning}
L.~Oneto, M.~Donini, A.~Maurer, and M.~Pontil.
\newblock Learning fair and transferable representations.
\newblock {\em CoRR}, abs/1906.10673, 2019.

\bibitem{oneto2019general}
L.~Oneto, M.~Donini, and M.~Pontil.
\newblock General fair empirical risk minimization.
\newblock {\em CoRR}, abs/1901.10080, 2019.

\bibitem{OnetoJ015}
L.~Oneto, S.~Ridella, and D.~Anguita.
\newblock Tikhonov, ivanov and morozov regularization for support vector
  machine learning.
\newblock {\em Machine Learning}, 103(1):103--136, 2015.

\bibitem{oneto2017dropout}
L.~Oneto, A.~Siri, G.~Luria, and D.~Anguita.
\newblock Dropout prediction at university of genoa: a privacy preserving data
  driven approach.
\newblock In {\em European Symposium on Artificial Neural Networks,
  Computational Intelligence and Machine Learning}, 2017.

\bibitem{papamitsiou2014learning}
Z.~Papamitsiou and A.~A. Economides.
\newblock Learning analytics and educational data mining in practice: A
  systematic literature review of empirical evidence.
\newblock {\em Journal of Educational Technology \& Society}, 17(4):49--64,
  2014.

\bibitem{pearl2000causality}
J.~Pearl.
\newblock {\em Causality: Models, Reasoning and Inference}.
\newblock Springer, 2000.

\bibitem{pearl2016causal}
J.~Pearl, M.~Glymour, and N.~P. Jewell.
\newblock {\em Causal Inference in Statistics: A Primer}.
\newblock John Wiley \& Sons, 2016.

\bibitem{pedreschi2009measuring}
D.~Pedreschi, S.~Ruggieri, and F.~Turini.
\newblock Measuring discrimination in socially-sensitive decision records.
\newblock In {\em SIAM International Conference on Data Mining}, 2009.

\bibitem{pedreshi2008discrimination}
D.~Pedreshi, S.~Ruggieri, and F.~Turini.
\newblock Discrimination-aware data mining.
\newblock In {\em ACM SIGKDD international conference on Knowledge discovery
  and data mining}, 2008.

\bibitem{perez-suay2017fair}
A.~P{\'e}rez-Suay, V.~Laparra, G.~Mateo-Garc{\'\i}a, J.~Mu{\~{n}}oz-Mar{\'\i},
  L.~G{\'o}mez-Chova, and G.~Camps-Valls.
\newblock Fair kernel learning.
\newblock In {\em Machine Learning and Knowledge Discovery in Databases}, 2017.

\bibitem{perlich2014machine}
C.~Perlich, B.~Dalessandro, T.~Raeder, O.~Stitelman, and F.~Provost.
\newblock Machine learning for targeted display advertising: Transfer learning
  in action.
\newblock {\em Machine Learning}, 95(1):103--127, 2014.

\bibitem{peters2017elements}
J.~Peters, D.~Janzing, and B.~Sch{\"o}lkopf.
\newblock {\em Elements of causal inference: foundations and learning
  algorithms}.
\newblock MIT press, 2017.

\bibitem{peyre2019computational}
M.~Peyr{\'e}, G.and M.~Cuturi.
\newblock Computational optimal transport.
\newblock {\em Foundations and Trends in Machine Learning}, 11(5-6):355--607,
  2019.

\bibitem{pleiss2017fairness}
G.~Pleiss, M.~Raghavan, F.~Wu, J.~Kleinberg, and K.~Q. Weinberger.
\newblock On fairness and calibration.
\newblock In {\em Proceedings of the 31st Conference on Neural Information
  Processing Systems}, pages 5684--5693, 2017.

\bibitem{quadrianto2017recycling}
N.~Quadrianto and V.~Sharmanska.
\newblock Recycling privileged learning and distribution matching for fairness.
\newblock In {\em Proceedings of the 31st Conference on Neural Information
  Processing Systems}, pages 677--688, 2017.

\bibitem{quionero2009dataset}
J.~Quionero-Candela, M.~Sugiyama, A.~Schwaighofer, and N.~D. Lawrence.
\newblock {\em Dataset Shift in Machine Learning}.
\newblock The MIT Press, 2009.

\bibitem{raff2018fair}
E.~Raff, J.~Sylvester, and S.~Mills.
\newblock Fair forests: Regularized tree induction to minimize model bias.
\newblock In {\em Proceedings of the 2018 AAAI/ACM Conference on AI, Ethics,
  and Society}, 2018.

\bibitem{redmond2009communities}
M.~Redmond.
\newblock {Communities and Crime Data Set}.
\newblock Available at
  \url{http://archive.ics.uci.edu/ml/datasets/communities+and+crime}, 2009.

\bibitem{rosenberg2018immigration}
M.~Rosenberg and R.~Levinson.
\newblock Trump's catch-and-detain policy snares many who call the {U.S.} home,
  2018.

\bibitem{russell2017worlds}
C.~Russell, M.~J. Kusner, J.~Loftus, and R.~Silva.
\newblock When worlds collide: integrating different counterfactual assumptions
  in fairness.
\newblock In {\em Proceedings of the 31st Conference on Neural Information
  Processing Systems}, pages 6414--6423, 2017.

\bibitem{selbst2017disparate}
A.~D. Selbst.
\newblock Disparate impact in big data policing.
\newblock {\em Georgia Law Review}, 52:109--195, 2017.

\bibitem{shalev2014understanding}
S.~Shalev-Shwartz and S.~Ben-David.
\newblock {\em Understanding Machine Learning: From Theory to Algorithms}.
\newblock Cambridge University Press, 2014.

\bibitem{shawe2004kernel}
J.~Shawe-Taylor and N.~Cristianini.
\newblock {\em Kernel Methods for Pattern Analysis}.
\newblock Cambridge University Press, 2004.

\bibitem{smola2001learning}
A.~J. Smola and B.~Sch{\"o}lkopf.
\newblock {\em Learning with Kernels}.
\newblock MIT Press, 2001.

\bibitem{song2018learning}
J.~Song, P.~Kalluri, A.~Grover, S.~Zhao, and S.~Ermon.
\newblock Learning controllable fair representations.
\newblock {\em CoRR}, abs/1812.04218, 2018.

\bibitem{speicher2018unified}
T.~Speicher, H.~Heidari, N.~Grgic-Hlaca, K.~P. Gummadi, A.~Singla, A.~Weller,
  and M.~B. Zafar.
\newblock A unified approach to quantifying algorithmic unfairness: Measuring
  individual \&group unfairness via inequality indices.
\newblock In {\em ACM SIGKDD International Conference on Knowledge Discovery \&
  Data Mining}, 2018.

\bibitem{spirtes2000causation}
P.~Spirtes, C.~N. Glymour, R.~Scheines, D.~Heckerman, C.~Meek, G.~Cooper, and
  T.~Richardson.
\newblock {\em Causation, Prediction, and Search}.
\newblock MIT press, 2000.

\bibitem{srebro2004learning}
N.~Srebro.
\newblock Learning with matrix factorizations, 2004.

\bibitem{stevenson2017assessing}
M.~T. Stevenson.
\newblock Assessing risk assessment in action.
\newblock {\em Minnesota Law Review}, 103, 2017.

\bibitem{strack2014diabetes}
B.~Strack, J.~P. DeShazo, C.~Gennings, J.~L. Olmo, S.~Ventura, K.~J. Cios, and
  J.~N. Clore.
\newblock {Diabetes 130-US hospitals for years 1999-2008 Data Set}.
\newblock Available at
  \url{https://archive.ics.uci.edu/ml/datasets/Diabetes+130-US+hospitals+for+years+1999-2008},
  2014.

\bibitem{OnetoC037}
M.~Vahdat, L.~Oneto, D.~Anguita, M.~Funk, and M.~Rauterberg.
\newblock A learning analytics approach to correlate the academic achievements
  of students with interaction data from an educational simulator.
\newblock In {\em European Conference on Technology Enhanced Learning}, 2015.

\bibitem{vaithianathan2013child}
R.~Vaithianathan, T.~Maloney, E.~Putnam-Hornstein, and N.~Jiang.
\newblock Children in the public benefit system at risk of maltreatment:
  Identification via predictive modeling.
\newblock {\em American Journal of Preventive Medicine}, 45(3):354--359, 2013.

\bibitem{verma2018fairness}
S.~Verma and J.~Rubin.
\newblock Fairness definitions explained.
\newblock In {\em IEEE/ACM International Workshop on Software Fairness}, 2018.

\bibitem{villani2009optimal}
C.~Villani.
\newblock {\em Optimal Transport Old and New}.
\newblock Springer, 2009.

\bibitem{wang2018invariant}
Y.~Wang, T.~Koike-Akino, and D.~Erdogmus.
\newblock Invariant representations from adversarially censored autoencoders.
\newblock {\em CoRR}, abs/1805.08097, 2018.

\bibitem{wightman1998law}
L.~F. Wightman.
\newblock {Law School Admissions}.
\newblock Available at \url{https://www.lsac.org/data-research}, 1998.

\bibitem{williamson2019fairness}
R.~C. Williamson and A.~K. Menon.
\newblock Fairness risk measures.
\newblock {\em Proceedings of the 36th International Conference on Machine
  Learning}, pages 6786--6797, 2019.

\bibitem{woodworth2017learning}
B.~Woodworth, S.~Gunasekar, M.~I. Ohannessian, and N.~Srebro.
\newblock Learning non-discriminatory predictors.
\newblock In {\em Computational Learning Theory}, 2017.

\bibitem{wu2016using}
Y.~Wu and X.~Wu.
\newblock Using loglinear model for discrimination discovery and prevention.
\newblock In {\em IEEE International Conference on Data Science and Advanced
  Analytics}, 2016.

\bibitem{yang2017measuring}
K.~Yang and J.~Stoyanovich.
\newblock Measuring fairness in ranked outputs.
\newblock In {\em International Conference on Scientific and Statistical
  Database Management}, 2017.

\bibitem{yao2017beyond}
S.~Yao and B.~Huang.
\newblock Beyond parity: Fairness objectives for collaborative filtering.
\newblock In {\em Proceedings of the 31st Conference on Neural Information
  Processing Systems}, pages 2921--2930, 2017.

\bibitem{yeh2016default}
I.~C. Yeh and C.~H. Lien.
\newblock {Default of Credit Card Clients Data Set}.
\newblock Available at
  \url{https://archive.ics.uci.edu/ml/datasets/default+of+credit+card+clients},
  2016.

\bibitem{yona2018probably}
G.~Yona and G.~Rothblum.
\newblock Probably approximately metric-fair learning.
\newblock In {\em Proceedings of the 35th International Conference on Machine
  Learning}, pages 5680--5688, 2018.

\bibitem{zafar2017fairness}
M.~B. Zafar, I.~Valera, M.~Gomez~Rodriguez, and K.~P. Gummadi.
\newblock Fairness beyond disparate treatment \& disparate impact: Learning
  classification without disparate mistreatment.
\newblock In {\em International Conference on World Wide Web}, 2017.

\bibitem{zafar2017fairnessARXIV}
M.~B. Zafar, I.~Valera, M.~Gomez~Rodriguez, and K.~P. Gummadi.
\newblock Fairness constraints: Mechanisms for fair classification.
\newblock In {\em Proceedings of the 20th International Conference on
  Artificial Intelligence and Statistics}, pages 962--970, 2017.

\bibitem{zafar2019fairness}
M.~B. Zafar, I.~Valera, M.~Gomez-Rodriguez, and K.~P. Gummadi.
\newblock Fairness constraints: A flexible approach for fair classification.
\newblock {\em Journal of Machine Learning Research}, 20(75):1--42, 2019.

\bibitem{zafar2017parity}
M.~B. Zafar, I.~Valera, M.~Rodriguez, K.~Gummadi, and A.~Weller.
\newblock From parity to preference-based notions of fairness in
  classification.
\newblock In {\em Proceedings of the 31st Conference on Neural Information
  Processing Systems}, pages 229--239, 2017.

\bibitem{zehlike2017matching}
M.~Zehlike, P.~Hacker, and E.~Wiedemann.
\newblock Matching code and law: Achieving algorithmic fairness with optimal
  transport.
\newblock {\em arXiv preprint arXiv:1712.07924}, 2017.

\bibitem{zemel2013learning}
R.~Zemel, Y.~Wu, K.~Swersky, T.~Pitassi, and C.~Dwork.
\newblock Learning fair representations.
\newblock In {\em Proceedings of the 30th International Conference on Machine
  Learning}, pages 325--333, 2013.

\bibitem{zhang2018mitigating}
B.~H. Zhang, B.~Lemoine, and M.~Mitchell.
\newblock Mitigating unwanted biases with adversarial learning.
\newblock In {\em Proceedings of the 2018 AAAI/ACM Conference on AI, Ethics,
  and Society}, pages 335--340, 2018.

\bibitem{zhang2017achieving}
L.~Zhang, Y.~Wu, and X.~Wu.
\newblock Achieving non-discrimination in data release.
\newblock In {\em ACM SIGKDD International Conference on Knowledge Discovery
  and Data Mining}, 2017.

\bibitem{zhang2017causal}
L.~Zhang, Y.~Wu, and X.~Wu.
\newblock A causal framework for discovering and removing direct and indirect
  discrimination.
\newblock In {\em Proceedings of the Twenty-Sixth International Joint
  Conference on Artificial Intelligence}, pages 3929--3935, 2017.

\bibitem{zhang2018achieving}
L.~Zhang, Y.~Wu, and X.~Wu.
\newblock Achieving non-discrimination in prediction.
\newblock In {\em Proceedings of the Twenty-Seventh International Joint
  Conference on Artificial Intelligence}, pages 3097--3103, 2018.

\bibitem{vzliobaite2011handling}
I.~Zliobaite, F.~Kamiran, and T.~Calders.
\newblock Handling conditional discrimination.
\newblock In {\em IEEE International Conference on Data Mining}, 2011.

\end{thebibliography}
\end{document}